\documentclass[journal,compsoc]{IEEEtran}
\IEEEoverridecommandlockouts

\usepackage{amsfonts}
\usepackage{amssymb}
\usepackage{array, makecell}
\usepackage{mathtools}
\usepackage{multirow}
\usepackage{amsthm}

\usepackage{graphicx}
\usepackage{subfigure}
\usepackage{enumerate}
\usepackage{amsmath}
\usepackage{color}
\usepackage{amsthm}
\usepackage{algorithm}
\usepackage[noend]{algpseudocode}
\usepackage{stfloats}

\usepackage{cite}

\theoremstyle{definition}

\newcommand{\bp}{\begin{proof} \small }
\newcommand{\ep}{\end{proof} \normalsize}
\newcommand{\epx}{\end{proof} \small}
\newcommand{\bpa}{\begin{proofappx} \footnotesize }
\newcommand{\epa}{\end{proofappx} \small }
\newtheorem{theorem}{Theorem}
\newtheorem{proposition}{Proposition}

\newtheorem{lemma}{Lemma}

\newtheorem*{theorem*}{Theorem}
\newtheorem*{proposition*}{Proposition}
\newtheorem*{corollary*}{Corollary}
\newtheorem*{lemma*}{Lemma}
\newtheorem*{assumption*}{Assumption}
\newtheorem*{definition*}{Definition}
\newtheorem*{claim*}{Claim}

\newcommand{\be}{\begin{equation}}
\newcommand{\ee}{\end{equation}}
\newcommand{\bs}{\begin{subequations}}
\newcommand{\es}{\end{subequations}}
\newcommand{\bq}{\begin{eqnarray}}
\newcommand{\eq}{\end{eqnarray}}
\newcommand{\bqn}{\begin{eqnarray*}}
\newcommand{\eqn}{\end{eqnarray*}}

\newcommand{\ba}{\left[ \begin{array}}
\newcommand{\ea}{\\ \end{array} \right]}
\newcommand{\ben}{\begin{enumerate}}
\newcommand{\een}{\end{enumerate}}

\def\real{{\mathchoice%
{\hbox{\rm\setbox1=\hbox{I}\copy1\kern-.45\wd1 R}}
{\hbox{\rm\setbox1=\hbox{I}\copy1\kern-.45\wd1 R}}
{\hbox{\scriptsize\rm\setbox1=\hbox{I}\copy1\kern-.45\wd1 R}}
{\hbox{\scriptsize\rm\setbox1=\hbox{I}\copy1\kern-.45\wd1 R}}}}

\def\Zint{{\mathchoice{\setbox1=\hbox{\sf Z}\copy1\kern-.75\wd1\box1}
{\setbox1=\hbox{\sf Z}\copy1\kern-.75\wd1\box1}
{\setbox1=\hbox{\scriptsize\sf Z}\copy1\kern-.75\wd1\box1}
{\setbox1=\hbox{\scriptsize\sf Z}\copy1\kern-.75\wd1\box1}}}
\newcommand{\complex}{ \hbox{\rm C\kern-0.45em\rule[.07em]{.02em}{.58em}%
\kern 0.43em}}

\makeatletter
\newcommand{\algmargin}{\the\ALG@thistlm}
\makeatother
\newlength{\whilewidth}
\algdef{SE}[parWHILE]{parWhile}{EndparWhile}[1]
{\parbox[t]{\dimexpr\linewidth-\algmargin}{%
		\hangindent\whilewidth\strut\algorithmicwhile\ #1\ \algorithmicdo\strut}}{\algorithmicend\ \algorithmicwhile}%
\algnewcommand{\parState}[1]{\State%
	\parbox[t]{\dimexpr\linewidth-\algmargin}{\strut #1\strut}}

\ifodd 1

\else

\fi

\def\BibTeX{{\rm B\kern-.05em{\sc i\kern-.025em b}\kern-.08em
		T\kern-.1667em\lower.7ex\hbox{E}\kern-.125emX}}
		
\newcommand\reg{\textsc{Regret}}
\begin{document}	
		
\title{Learning the Optimal Path and DNN Partition for Collaborative Edge Inference}

\author{Yin Huang ~\IEEEmembership{Student Member,~IEEE}
       ~~Letian Zhang~\IEEEmembership{Member,~IEEE}
       ~~Jie~Xu~\IEEEmembership{Senior Member,~IEEE}
\IEEEcompsocitemizethanks{\IEEEcompsocthanksitem Y. Huang and J. Xu are with the Department of Electrical and Computer Engineering, University of Miami. Email: \{yxh954, jiexu\}@miami.edu.\IEEEcompsocthanksitem L. Zhang is with the Department of Computer Science, Middle Tennessee State University. Email: letian.zhang@mtsu.edu.
}}

\IEEEtitleabstractindextext{%
\begin{abstract}
Recent advancements in Deep Neural Networks (DNNs) have catalyzed the development of numerous intelligent mobile applications and services. However, they also introduce significant computational challenges for resource-constrained mobile devices. To address this, collaborative edge inference has been proposed. This method involves partitioning a DNN inference task into several subtasks and distributing these across multiple network nodes. Despite its potential, most current approaches presume known network parameters—like node processing speeds and link transmission rates—or rely on a fixed sequence of nodes for processing the DNN subtasks. In this paper, we tackle a more complex scenario where network parameters are unknown and must be learned, and multiple network paths are available for distributing inference tasks. Specifically, we explore the learning problem of selecting the optimal network path and assigning DNN layers to nodes along this path, considering potential security threats and the costs of switching paths. We begin by deriving structural insights from the DNN layer assignment with complete network information, which narrows down the decision space and provides crucial understanding of optimal assignments. We then cast the learning problem with incomplete network information as a novel adversarial group linear bandits problem with switching costs, featuring rewards generation through a combined stochastic and adversarial process. We introduce a new bandit algorithm, B-EXPUCB, which combines elements of the classical blocked EXP3 and LinUCB algorithms, and demonstrate its sublinear regret. Extensive simulations confirm B-EXPUCB's superior performance in learning for collaborative edge inference over existing algorithms.

\end{abstract}

}

\maketitle
\IEEEdisplaynontitleabstractindextext
\IEEEpeerreviewmaketitle
\IEEEraisesectionheading{\section{Introduction}}
\IEEEPARstart{A} growing number of contemporary applications and services, such as augmented reality/virtual reality, face recognition, and speech assistant, demand real-time inferencing of deep neural networks (DNNs), which can effectively extract high-level features from the raw data at high computational complexity cost~\cite{chiang2016fog,sze2017efficient}.  In many cases of
practical importance, these applications often are run at resource-constrained embedded devices,
such as mobile phones, wearables and IoT devices
in next-generation networks~\cite{gubbi2013internet,yang2020artificial}. However, it is very challenging to compute the DNN inference tasks at resource-constrained devices due to the limited computation and battery capacity.
Therefore, it is crucial to leverage external computation resources to realize the full potential and benefits of future device-based artificial intelligence (AI) applications.

Collaborative edge inference presents an effective method for executing DNN inference, overcoming the limitations of resource-constrained devices by leveraging resources available on nearby devices at the network edge. Its fundamental concept involves partitioning a DNN into multiple segments, each comprising a subset of the DNN layers. This division allows for the inference task to be split into multiple sub-tasks that can be distributed across multiple devices. By combining the capabilities of both on-device processing and computation offloading, resource-constrained devices can delegate all or part of their inference workload to network edge devices, such as cellular base stations, access points, or peer devices within the same network. With the advancement of collaborative edge inference technologies, the real-time computation of complex DNNs on resource-constrained devices becomes feasible.




\begin{figure}[tb]
	\centering
	\includegraphics[width=1\linewidth]{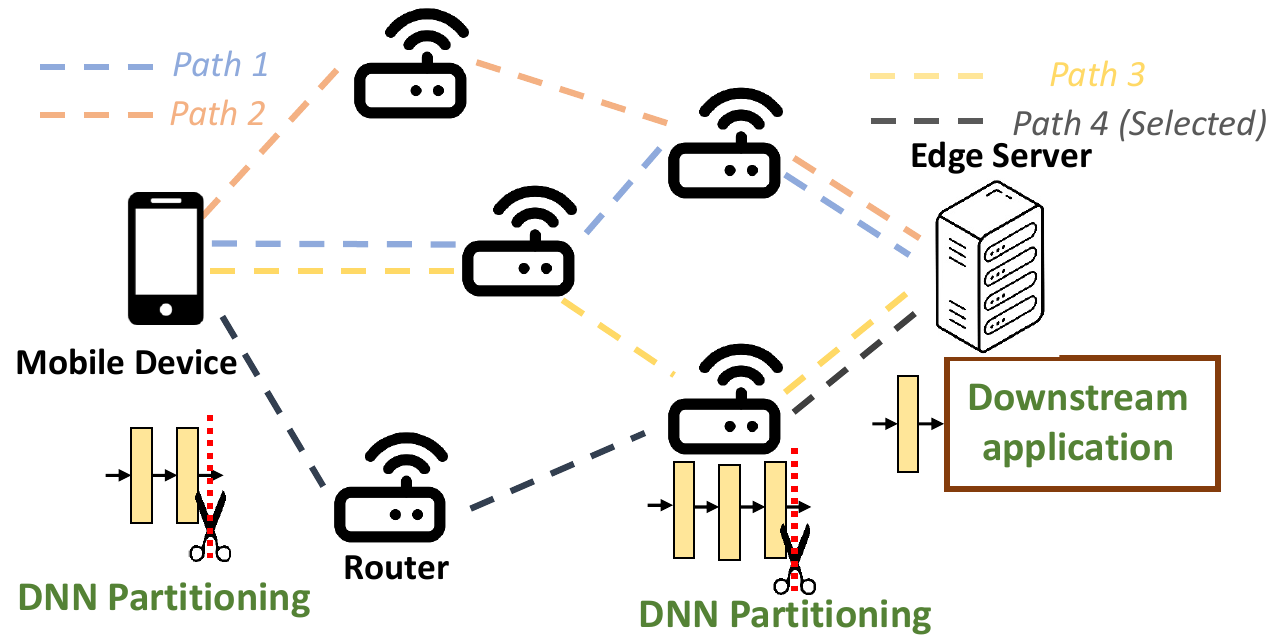}
	\caption{An illustration of collaborative edge inference.} \label{fig:collaborative_edge_inference}
\end{figure}%

Despite the promise of collaborative edge inference, the majority of existing efforts have been predominantly focusing on partitioning the DNN between two devices~\cite{kang2017neurosurgeon, eshratifar2019jointdnn,wang2023sequential,hu2019dynamic,zhang2021autodidactic,li2019edge}, typically a resource-constrained mobile device and a powerful edge server. However, these studies overlook the potential benefits of leveraging multiple devices, such as peer devices and relay devices, within real-world communication networks to further optimize the collaborative edge inference performance. On the one hand, these additional devices can offer supplementary spare computing resources, facilitating the execution of more complex DNN inference tasks. On the other hand, these devices can also provide alternative paths between the mobile device and the edge server, particularly useful when the direct communication link experiences poor conditions. An illustration of collaborative edge inference over multiple devices is shown in Fig.~\ref{fig:collaborative_edge_inference}.  However, fully realizing the potential of collaborative edge inference across multiple devices entails addressing several key challenges. \textbf{First}, compared to partitioning the DNN between just two devices, the decision space expands significantly. This expansion involves not only selecting an appropriate communication path between the mobile device and the edge server but also assigning DNN layers to the devices along the path to distribute the DNN inference workload effectively. \textbf{Second}, the end-to-end inference latency is contingent on the computing speed of the devices and the transmission speed of the links along the selected path. These speeds fluctuate over time and may not be known beforehand when making collaborative inference decisions. Furthermore, potential attacks on both the devices and the links can introduce additional uncertainty to this information. \textbf{Third}, switching devices for collaborative edge inference necessitates reloading and setting up the DNN models in the device memory, which introduces overhead and can impact the overall inference latency. Thus, frequent device and path switching is undesirable.

In this paper, we study the joint optimization of path selection and DNN layer assignment in collaborative edge inference across multiple devices. Specifically, we explore scenarios where multiple paths exist between a source node (i.e., the mobile device) and a destination node (i.e., the edge server), with a focus on optimizing path selection and DNN layer assignments to minimize end-to-end inference delay. Addressing challenges stemming from information uncertainty, potential attacks, and switching overhead, we introduce a novel adversarial group linear bandits algorithm designed to dynamically learn optimal path selection and DNN layer assignment over time. Our primary contributions can be outlined as follows.

\begin{itemize}
\item We formulate the joint path selection and DNN layer assignment problem in collaborative edge inference across multiple devices, with the objective of minimizing end-to-end inference latency. We analyze the structural properties of optimal DNN layer assignment given complete information regarding node computing speed and link transmission speed.

\item Given the absence of prior information on node computing speed and link transmission speed, we cast the joint path selection and DNN layer assignment problem as a novel adversarial group linear bandits problem. Introducing a new bandit algorithm, termed B-EXPUCB, we aim to learn the optimal path and DNN layer assignment over time. Our formulation treats candidate paths as groups and the potential DNN layer assignments along each path as arms within the group. 

\item Given that arms within the same group share common linear parameters, namely the unknown node computing speed and link transmission speed, learning within a group manifests as a stochastic linear bandit problem. Moreover, since the end-to-end inference performance may be susceptible to potential attacks, the optimal path selection represents an adversarial bandit problem. To address these challenges, our algorithm integrates classical stochastic linear bandit techniques such as LinUCB with adversarial bandit strategies like EXP3 in a cohesive manner. Additionally, we incorporate a block structure to alleviate switching costs. Through careful parameter selection, we establish a reward regret bound of $O(T^{3/4} \sqrt{\log T})$ with high probability and a switching cost bound of $O(T^{1/3})$.

\item We conduct comprehensive simulations studies to demonstrate the superiority of the proposed B-EXPUCB algorithm compared to existing bandit algorithms in optimizing end-to-end collaborative edge inference performance.
\end{itemize}

\begin{table*}[]
\centering
\caption{Comparison of Collaborative Edge Inference Approaches}
\begin{tabular}{|c|c|c|c|c|c|c|c|}
\hline
                     & ~\cite{zhang2021autodidactic,li2019edge} & ~\cite{kang2017neurosurgeon, eshratifar2019jointdnn,wang2023sequential,hu2019dynamic} & ~\cite{he2020joint} & ~\cite{zhou2019distributing,mohammed2020distributed,zeng2020coedge,xu2022distributed} & ~\cite{hou2022distredge} & ~\cite{cui2022learning} & Ours           \\ \hline
single/multi hop     & single hop                                             & single hop                                                                                    & single hop                                & multi hops                                                                                                  & multi hops                                     & single hop                                    & multi hops     \\ \hline
online/offline       & online                                              & offline                                                                                       & offline                                   & offline                                                                                                     & online                                         & online                                        & online         \\ \hline
fixed/path selection & fixed path                                          & fixed path                                                                                    & path selection                            & single path                                                                                                 & single path                                    & path selection                                & path selection \\ \hline
attack/no attack     & no attack                                           & no attack                                                                                     & no attack                                 & no attack                                                                                                   & no attack                                      & no attack                                     & attack         \\ \hline
\end{tabular}
\end{table*}

\section{Related Work}

\textbf{Collaborative Edge Inference}. 
We present a comparison with existing collaborative edge inference approaches, with a corresponding summary provided in Table I for clarity. Prior research on collaborative edge inference has explored DNN partitioning optimization in both offline settings ~\cite{kang2017neurosurgeon, eshratifar2019jointdnn,wang2023sequential,hu2019dynamic} and online settings ~\cite{li2019edge,zhang2021autodidactic}. For example, in~\cite{zhang2021autodidactic}, the authors treat potential partition points as arms and consider the transmission rate between mobile devices and servers, along with server computing speed, as parameters to be learned. However, these efforts mainly focus on collaboration between two devices. Some studies have investigated collaborative edge inference between a mobile device and multiple edge servers in both offline~\cite{he2020joint} and online settings ~\cite{cui2022learning}. In~\cite{cui2022learning}, for instance, the authors propose a deep reinforcement learning approach to address DNN inference task offloading in queue-based multi-device and multi-server collaborative edge computing. Yet, these works primarily examine DNN partitioning on single-hop paths between the source and destination nodes.

Several other works ~\cite{zhou2019distributing,mohammed2020distributed,zeng2020coedge,xu2022distributed, hou2022distredge}  consider collaborative edge inference in multi-hop scenarios, leveraging nearby idle devices' computing resources. For example, authors of ~\cite{mohammed2020distributed} propose a fine-grained adaptive partitioning scheme to divide DNNs into multiple partitions processed locally by end devices and nearby powerful nodes. Nonetheless, these studies typically focus on pre-determined single paths and do not delve into path selection in collaborative edge inference.

It is important to note numerous works addressing the broader computation offloading problem ~\cite{sardellitti2015joint,lyu2016multiuser,tran2018joint, du2019service,ouyang2019adaptive,wang2023online}. Some studies have also considered the presence of attackers and proposed risk-aware offloading strategies in multi-server offloading scenarios ~\cite{apostolopoulos2020risk, bai2020risk}. However, these investigations do not specifically tackle the challenge of DNN partitioning.

\textbf{Multi-armed Bandits.} Bandit problems are typically categorized as either stochastic bandits or adversarial bandits based on how rewards are generated~~\cite{bubeck2012regret}. The classical UCB algorithm~\cite{auer2002finite} and EXP3 algorithm~\cite{auer2002nonstochastic} have been developed for optimal regret bounds in stochastic and adversarial bandits, respectively. 
Recent endeavors has exploited the intersection of stochastic and adversarial bandits. In one line of work~\cite{bubeck2012best,seldin2014one}, efforts have been made to design a single algorithm applicable to both regimes without prior knowledge. However, these works consider rewards from a single distribution. In another line of research~\cite{lykouris2018stochastic,gupta2019better}, the focus is on stochastic bandits with adversarial corruption of reward observations, while the actual received reward remains unaffected.

Another important extension considers the inclusion of switching costs~\cite{agrawal1988asymptotically}, which arise when changing actions between consecutive rounds incurs non-negligible overhead costs. Addressing bandit problems with switching costs often entails segmenting time into progressively longer blocks. 
Recent state-of-the-art block-based algorithms~\cite{rouyer2021algorithm,amir2022better} have achieved a regret of $O(T^{2/3})$ in the more general adversarial bandits setting, matching the minimax lower bound, and a regret of $O(T^{1/3})$ in the stochastic bandits setting. 

Our previous work~\cite{huang2023adversarial} considers joint rewards influenced by both stochastic distribution and adversarial behavior.  A new bandit algorithm is developed and applied to address server selection in collaborative edge inference. However, collaborative edge inference is limited to single-hop paths, and switching costs are not considered. In this present work, we investigate the general multi-hop collaborative edge inference problem while considering path switching costs. Additionally, we characterize the structural properties of the optimal DNN layer assignment.


\section{System Model}
\subsection{Problem Formulation}

We consider a collaborative inference problem for a resource-constrained mobile device. In addition to the edge server, there are other available devices in the network, functioning as relays and creating multiple possible communication paths between the mobile device and the edge server. The mobile device needs to sequentially process a series of DNN inference tasks, denoted by the task index $t= 1,\dots,T$. Each task $t$ is a DNN inference task using DNN$^t$, which might change over time or remain constant. 
Each inference task utilizes input data collected by the mobile device and aims to deliver the inference results to the edge server for downstream applications/services. Due to the layered structure of DNNs, the mobile device has the flexibility to divide each task into several distinct parts and can leverage the relays or the server to assist inference by running these parts of DNN on these devices. Define a set $\mathcal{L}^t = \{1, 2, ..., L^t\}$, representing layer index of DNN$^t$, where $L^t$ is the number of layers of DNN$^t$. We use $c_l$ to denote the computational workload of layer $l$ and the $s_l$ to denote the output data size of layer $l$, $\forall l \in \mathcal{L}^t$. Therefore, $s_{L^t}$ is the final output data size. In addition, we use $s_0$ to denote the input raw data size. 

The communication network linking a mobile device with an edge server can be represented by an acyclic directed graph, denoted as $\langle \mathcal{V}, \mathcal{E} \rangle$. Here, $\mathcal{V}$ is the set of nodes, which includes the mobile device, any potential relays, and the edge server, while $\mathcal{E}$ denotes the set of links between these nodes. Each link $e = (u, v) \in \mathcal{E}$ represents a communication channel where data travels from node $u$ to node $v$. The computational speed of any node $v$ is given by $p_v$, and the transmission rate of a link $e = (u, v)$ is represented by $b_e$. Additionally, $\mathcal{G}$ is a set representing $G$ possible paths from the mobile device to the server. Each path $g \in \mathcal{G}$ consists of a series of links and nodes from $\mathcal{E}$ and $\mathcal{V}$ that create a contiguous route from the mobile device to the edge server.

For a specific path $g \in \mathcal{G}$, the node set and link set are represented by $\mathcal{V}_g \subset \mathcal{V}$ and $\mathcal{E}_g \subset \mathcal{E}$, respectively. The assignment of DNN layers to nodes along this path is determined by a DNN layer assignment function $f_g$, which maps the layers to nodes: $f_g: \mathcal{L}^t \rightarrow \mathcal{V}_g$. This means that if $f_g(l) = v$, then layer $l$ of the DNN is processed by node $v$. The layer assignment function $f_g$ adheres to two key properties for collaborative inference:
\begin{itemize}
    \item \textbf{Many-to-One}: Multiple consecutive DNN layers may be processed by a single node.
    \item \textbf{Non-decreasing}: If a layer $l$ is assigned to a node $v$, subsequent layers cannot be assigned to any nodes preceding $v$ in the path.
\end{itemize}

The total inference time for implementing an assignment $f_g$ on path $g$, which includes both the processing time of the layers and the data transmission time, is calculated as:
\begin{align}
    d(g,f_g) = &\sum_{i < f_g(1)} \frac{s_0}{b_i} + \sum_{l=1}^{L-1}\left(\frac{c_l}{p_{f_g(l)}} + \sum_{f_g(l) < i < f_g(l+1)}\frac{s_l}{b_i}\right) \nonumber
    \\&+ \frac{c_L}{p_{f_g(L)}} + \sum_{i > f_g(L)} \frac{s_{L}}{b_i},
    \label{eq:layer_a}
\end{align}
where the notation $i < v$ and $i > v$ denotes the links before and after node $v$ on the path, respectively. The term $\frac{s_l}{b_i}$ represents the time required for link $i$ to transmit the output of layer $l$, and $\frac{c_l}{p_{f_g(l)}}$ denotes the processing time required for node $f_g(l)$ to process layer $l$. Note that layer assignment encompasses two traditional approaches as \textbf{special cases}: (1) the mobile device carries out the complete inference task, which may require significant time, before transmitting the results to the edge server; (2) the mobile device sends the raw input data, which may also be time-consuming, to the edge server that then handles the full inference task.

Besides the inherent uncertainties stemming from variations in computing speed and transmission rates, the collaborative edge inference system may encounter additional non-stochastic uncertainties. We consider the potential presence of an adversary in the system, which aims to select paths to attack during each time slot. For instance, it may be susceptible to deliberate jamming attacks on the communication channel. Such attacks could lead to significant delays in processing a task. Alternatively, the channel might face substantial interference from concurrent transmissions in the vicinity, even in the absence of adversarial intent. In such situations, the inference delay can become prohibitively long, effectively reducing the benefit of collaborative inference to 0.

We examine a learning scenario where a learner (specifically, a mobile device) initially lacks knowledge of the node processing speeds and the link transmission speeds, and faces potential adversarial attacks. For each task $t$, the mobile device must chooses a path $g^t$ and a corresponding DNN layer assignment $f_{g^t}$ for collaborative DNN inference. Simultaneously, an adversary determines a binary attack vector $a^t = (a^t(g), \forall g\in \mathcal{G})$, where $a^t (g) = 0$ indicates an attack on the path $g$ and $a^t(g)=1$ indicates no attack. Given an inference deadline $D$, the reward is calculated based on the time difference between total inference process and the deadline. The reward function for task $t$ is therefore influenced by the chosen path, the DNN layer assignment, and the attacks, and is defined as follows:
\begin{align}
    r^t(g^t,f_{g^t}) = a^t(g^t)(D - d(g^t,f_{g^t}))
\end{align}

In this setting, we also account for a switching cost when changing selected paths. For instance, if the path chosen by the learner differs from the one used for the previous task, additional costs due to reloading and setting up the DNN on the nodes of the new path are incurred. The switching cost function is defined as $Q(g^t, g^{t-1}) = {\tau}\mathbf{1}\{g^t \neq g^{t-1}\}$, where $\tau$ represents the cost associated with switching paths.

\textbf{Objective}: The goal for the learner is to minimize the total inference delay while concurrently reducing the cumulative switching costs across $T$ inference tasks ($t = 1, ..., T$). This involves strategically choosing the optimal path and corresponding DNN layer assignment for each task to maximize the following objective function:
\begin{equation}
    \max \left[\sum_{t=1}^T [r^t(g^t,f^t) - Q(g^t,g^{t-1})]\right] .\label{obj_func}
\end{equation}
This scenario presents an online joint path selection and DNN layer assignment problem. On one hand, the learner must choose the optimal path considering the state of the communication network and striving to avoid attackers as effectively as possible. On the other hand, the learner must determine the best DNN layer assignment for the chosen path to minimize total inference delay. However, the learner does not initially know the processing speeds of the nodes or the transmission rates of the links, which must be learned in an online manner.

\section{Structural Results of Optimal Layer Assignment}
Before introducing the learning algorithm for determining the optimal path selection and DNN layer assignment, we first analyze the problem with complete information regarding node processing speed and link transmission speed. This initial analysis serves to simplify the learning problem when faced with incomplete information. Specifically, the potential DNN layer assignments over a path increase exponentially with the number of hops on that path. For a path with $N$ hops and a DNN comprising $L$ layers, the total number of possible assignments can be calculated as $\sum_{k=1}^{\min(N,L)}{N \choose k}{L \choose L+1-k}$. Consequently, as the number of hops on a path and the number of DNN layers increase, the decision space expands significantly. This section aims to develop structural insights into optimal layer assignment to mitigate the expansive decision space.

Our first result highlights the splitting positions within a DNN that would never emerge in the optimal layer assignment based solely on information about the DNN architecture. This means that even without knowledge of node processing speeds and link transmission speeds, and without solving the optimal layer assignment problem, it is possible to pre-group DNN layers into blocks. Consequently, rather than conducting DNN layer assignment directly, we can alternatively pursue DNN block assignment.

\begin{proposition}
    Assume that among all layer outputs and the input, the data size of the output layer is the smallest, i.e., $s_L < s_l, \forall l \in \{0, ..., L-1\}$. In the optimal layer assignment, the data sizes at the intermediate output splitting points form a non-increasing sequence.
\end{proposition}
\begin{proof}
Suppose in the optimal layer assignment, the DNN is partitioned into a number of $K$ block of layers, denoted by $B_1 \to B_2 \to ... \to B_K$. We use the notation $[k]$ to denote the last layer of block $B_k$, thus the data sizes at the intermediate output splitting points form a sequence $s_{[1]}, s_{[2]}, ..., s_{[K]}$. We also denote $v_{[k]}$ as the node to which block $B_k$ is assigned. 

Suppose the sequence of data sizes at the intermediate output splitting points is not non-increasing. Then, there must be splitting points $k'$ and $k''$, where $k' < k''$, so that $s_{[k']} < s_{[k'+1]} = ... = s_{[k'']} > s_{[k''+1]}$. A specific instance occurs when $k'+1 = k''$, simplifying to  $s_{[k''-1]} < s_{[k'']} > s_{[k''+1]}$. This condition arises if the sequence is not non-increasing, indicating an increase at some splitting point. Furthermore, since the smallest data size is at the final output by assumption, there must subsequently be a decrease at another splitting point. 

For all nodes $v_{[k'+1]}, ..., v_{[k'']}$, their processing speeds must be identical. If not, assigning all blocks $B_{k'+1}, ..., B_{k''}$ to the fastest among these nodes would reduce the processing time without increasing the transmission time, contradicting the optimal layer assignment assumption. 

Considering two scenarios, we can construct alternative layer assignments that reduce overall inference time:
\begin{itemize}
    \item Case where $p_{v_{[k'']}} \geq p_{v_{[k''+1]}}$: Merging blocks $B_{k''}$ and $B_{K''+1}$ into a single block and assign it to node $v_{[k'']}$ decreases the transmission time (since $s_{[k'']} > s_{[k''+1]}$) without increasing the processing time of these layers. 
    \item Case where $p_{v_{[k'']}} < p_{v_{[k''+1]}}$: Merging blocks $B_{k'+1}, ..., B_{k''}$ and $B_{k''+1}$ into a single block and assign it to node $v_{[k''+1]}$ decreases the transmission time (since $s_{[k']} < s_{[k'+1]}$) without increasing the processing time of these layers. 
\end{itemize}
This proves that if the sequence of data sizes at the intermediate output splitting points is not non-increasing, then the layer assignment must not be optimal. 
\end{proof}

\begin{figure}[tb]
	\centering
	\includegraphics[width=0.7\linewidth]{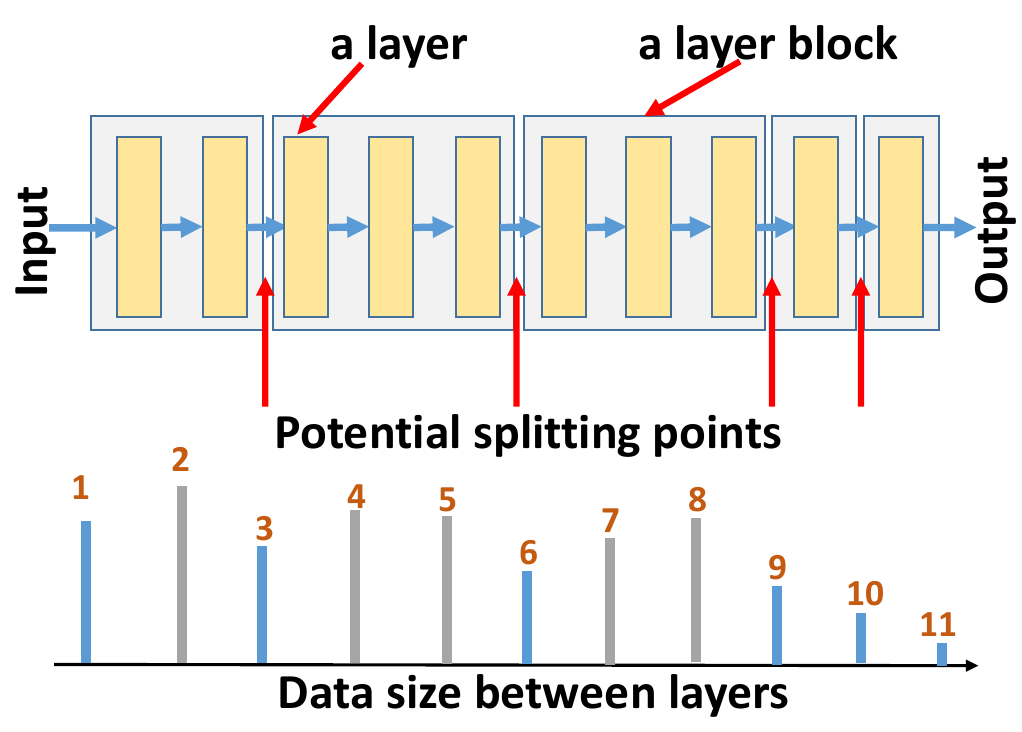}
	\caption{Identifying Potential Splitting Points of a DNN.} \label{fig:layers_block}
\end{figure}

Proposition 1 enables us to identify potential splitting points in an optimal layer assignment and to discard those that cannot be optimal. Specifically, a simple algorithm (see Algorithm \ref{splitting}) can be implemented by sequentially scanning through the layers. This algorithm retains splitting points only if they ensure that the sequence of layer output data sizes remains non-increasing. An example of this process is demonstrated in Fig.~\ref{fig:layers_block}, where 12 layers of a DNN are efficiently grouped into 5 ``super-layers'' using the algorithm. For the rest of this paper, we will assume that these potential splitting points have been identified, and we will refer to these super-layers simply as layers. Specifically, for a DNN consisting of $L$ layers, the data sizes satisfy $s_0 \geq s_1 \geq \ldots \geq s_{L}$.

\begin{algorithm}[t]
	\caption{Potential Splitting Point Identification}
	\begin{algorithmic}[1] 
		\State \textbf{Input}: The layer output data sizes of a DNN, $s_0, s_1, ..., s_L$. 
        \State \textbf{Output}: The potential splitting points $\mathcal{S}$
		\State \textbf{Initialization}: Set $\bar{s} = s_0$ and $\mathcal{S} = \emptyset$
        \For{$l = 1, ..., L-1$}
            \If{$s_l \leq \bar{s}$}
                \State Insert $l$ into $\mathcal{S}$
                \State Set $\bar{s} = s_l$
            \EndIf
        \EndFor
        \State Return $\mathcal{S}$
	\end{algorithmic}
	\label{splitting}
\end{algorithm}

Our second result highlights that, with full knowledge of the node processing speeds, it is possible to identify nodes that would not be assigned any layer processing tasks in the optimal layer assignment.
\begin{proposition}
In an optimal layer assignment, the processing speeds of nodes assigned with layer processing tasks form a non-decreasing sequence.
\end{proposition}
\begin{proof}
    Suppose in the optimal layer assignment, the DNN is partitioned into a number of $K$ block of layers, denoted by $B_1 \to B_2 \to ... \to B_K$. We use the notation $[k]$ to denote the last layer of block $B_k$. We also denote $v_{[k]}$ as the node to which block $B_k$ is assigned. According to Proposition 1, we have $s_0 \geq s_{[1]} \geq s_{[2]} \geq s_{[K]}$. 

    Suppose the sequence of $v_{[k]}$ is not non-decreasing, then there must exist some $k'$ so that $v_{[k']} > v_{[k'+1]}$. In this case, merging blocks $B_{[k']}$ and $B_{[k'+1]}$ into a single block and assigning it to $v_{[k']}$ would reduce the processing time without increasing the transmission time (since $s_{[k']} \geq s_{[k'+1]}$), contradicting the assumption of the optimal layer assignment assumption. 
\end{proof}
Proposition 2 allows us to determine which nodes along a path are assigned processing tasks, based on information about node processing speeds. A corresponding algorithm, following the same principles as Algorithm 1, can be implemented by sequentially scanning through the nodes along the path. This algorithm retains only those nodes for which the sequence of processing speeds forms a non-decreasing sequence.

We are also interested in understanding the conditions under which collaborative inference degenerates into the traditional single-node inference scenario. Proposition 2 provides insight into a specific situation: if the mobile device possesses the highest processing speed among all nodes along the path, then assigning the entire DNN to the mobile device is optimal. Proposition 3 extends this by offering a sufficient condition for more general scenarios, where any node might be the one tasked with processing the entire DNN.
\begin{proposition}
For a given path $g$, assigning all DNN layers to a node $v$ is optimal if: (1) $v$ is among the candidate nodes identified by Proposition 2. (2) The following conditions are met:
        \begin{align}
        \frac{p^{-1}_v - (\max_u p_u)^{-1}}{\sum_{v < i < v'} b^{-1}_i} &\leq \min_l c^{-1}_l(s_{l-1} - s_l)
        \label{sub}\\
        \frac{p^{-1}_{v''} - p^{-1}_v}{\sum_{0<i<v}b^{-1}_i} &\geq \max_l c^{-1}_l(s_{l-1} - s_l) \label{pre}
    \end{align}
Here, $v'$ and $v''$ denote the subsequent and preceding candidate nodes of $v$, respectively, identified by Proposition 2. 
\end{proposition}
\begin{proof}
Consider any assignment among the nodes identified by Proposition 2 (which may not be optimal). We examine the condition under which adding a layer to node $v$ would reduce the overall inference time. There are two scenarios to analyze: transferring a layer initially allocated to a subsequent node to node $v$ and transferring a layer originally assigned to a preceding node to node $v$.

Case 1: Let $\tilde{v}' \geq v'$ be the first subsequent node of $v$ assigned some DNN layers. Suppose the initial layer on node $v'$ is layer $l$. Moving layer $l$ to node $v$ for processing results in a change in the total inference time given by:
    \begin{align}
        &\frac{c_l}{p_v} + \sum_{v < i < \tilde{v}'}\frac{s_l}{b_i} - \left(\frac{c_l}{p_{\tilde{v}'}} + \sum_{v < i < \tilde{v}'}\frac{s_{l-1}}{b_i}\right) \\
        =& c_l (p^{-1}_{v} - p^{-1}_{\tilde{v}'}) - \sum_{v < i < \tilde{v}'} b^{-1}_i (s_{l-1} - s_{l}) \leq 0
    \end{align}
The inequality holds because of equation \eqref{sub}.

Case 2: Let $\tilde{v}'' \leq v''$ be the last preceding node of $v$ assigned some DNN layers. Suppose the last layer on node $\tilde{v}''$ is layer $l$. Transferring layer $l$ to node $v$ for processing leads to a change in the total inference time as follows:
    \begin{align}
        &\frac{c_l}{p_v} + \sum_{\tilde{v}'' < i < v} \frac{s_{l-1}}{b_i} - \left(\frac{c_l}{p_{\tilde{v}''}} + \sum_{\tilde{v}'' < i < v} \frac{s_l}{b_i}\right)\\
        =& \sum_{\tilde{v}'' < i < v} b^{-1}_i (s_{l-1} - s_{l}) - c_l(p^{-1}_{\tilde{v}''} - p^{-1}_v) \leq 0
    \end{align}
The inequality holds because of equation \eqref{pre}.
\end{proof}

\section{Learning Under Incomplete Information}
In this section, we delve into the collaborative inference problem under incomplete information and propose a novel bandits algorithm to learn optimal decisions regarding path selection and DNN layer assignment. We begin by framing the problem as an adversarial group linear bandits with switching cost scenario and then present our algorithm, termed B-EXPUCB. Following that, we establish the regret bound for our proposed algorithm.

\subsection{Adversarial Group Linear Bandits with Switching Cost}
We view the path selection and DNN layer assignment problem as a sequential decision-making process spanning $T$ rounds.
There are $G$ arm groups, denoted by $\mathcal{G} = \{1, 2, ..., G\}$, corresponding to the potential paths available. Each group $g \in \mathcal{G}$ is associated with an unknown parameter $\theta_g \in \mathbb{R}^{D_g}_+$, where $D_g = |\mathcal{V}_g| + |\mathcal{E}_g|$ represents the number of nodes and links along path $g$. Consequently, $\theta_g$ encompasses the unknown node processing speeds and link transmission speeds associated with group $g$.

A layer assignment $f_g$ along path $g$ yields a distribution of computational workload among the nodes and communication workload among the links, contingent on the DNN architecture in use. This distribution is denoted by a vector $x \in \mathbb{R}^{D_g}_+$. Let $\mathcal{X}^t_g$ denote the set of possible distribution vectors, corresponding to the potential layer assignments, for DNN task $t$ along path $g$. The total inference time when utilizing path $g$ and layer assignment $x$ is represented as $d(g, x) = \theta_g^\top x$. To align with reward terminology, we define $\tilde{\theta}_g = (1, -\theta_g)$ and $\tilde{x} = (D, x)$. Thus, the reward can be expressed as a linear function: $D - d(g, x) = \langle\tilde{\theta}_g, \tilde{x}\rangle$. For simplicity in notation, we use $\theta$ to denote $\tilde{\theta}$ and $x$ to denote $\tilde{x}$ without causing confusion.

Collaborative inference may also face potential attacks, as per our model. Thus, the ultimate reward of a decision regarding path selection and DNN layer assignment, denoted as $(g^t, x^t)$, is defined as follows:
\begin{align}
r^t(g^t, x^t, a^t) = a^t(g^t)\langle \theta_{g^t}, x^t\rangle
\end{align}
We operate under the assumption that the learner cannot directly observe the processing speeds of individual nodes or the transmission speeds of individual links for task $t$. Instead, it can only observe the realized reward of a task $t$ based on the chosen path and layer assignment.

\textbf{Optimal Benchmark}. As the adversary has the flexibility to employ various attacking strategies on the groups, our approach focuses on establishing an optimal benchmark based on hindsight among the group-static strategies. Specifically, a group-static strategy entails selecting a fixed group for the entire duration of $T$ rounds, although the chosen arm within that group can vary. In essence, for any given group $g$ in round $t$, it is possible to compute the optimal arm, denoted as $\xi^t_g \triangleq \arg\max_{x \in \mathcal{X}^t_g}\langle \theta_g, x\rangle$, to maximize the expected reward, irrespective of the adversary's attack strategy. In the special scenario where the set of available arms $\mathcal{X}^t_g$ remains consistent across all rounds, the optimal arm $\xi^t_g$ for group $g$ also remains static.

With the optimal arms in each group for each task $t$ understood, the optimal group given an attacking sequence $a^1, ..., a^T$ is thus the one that maximizes the total expected reward, which corresponds to minimizing the total inference delay,
\begin{align}
    \gamma(a^1, ..., a^T) = \arg\max_{g \in \mathcal{G}} \mathbb{E}\left[\sum_{t=1}^T r^t(g, \xi^t_g, a^t)\right]
    \label{optimal},
\end{align}
where the expectation is over the random noises. For notation simplicity, we also write $\gamma(a^1, ..., a^T) = \gamma$ by dropping the attack sequences but the readers should be cautious that the optimal group $\gamma$ depends on the attacks (and $T$). 

\textbf{Regret}. To optimize the objective function specified in \eqref{obj_func}, we use the reward regret and the switching regret to measure the performance of the algorithm. The reward regret of the learner is quantified as the difference between the expected cumulative reward achieved by the optimal benchmark and the cumulative reward obtained by the learner's algorithm,  
\begin{align}
    R_T = \mathbb{E}\left[\sum_{t=1}^T r^t(\gamma, \xi^t_\gamma) - \sum_{t=1}^T r^t(g^t, x^t)\right],
\end{align}
where the expectation is taken with respect to the noise and the possible internal randomization of the algorithm. 

The switching regret is defined as the accumulative switching cost up to $T$ round,
\begin{equation}
    S_T = \sum_{t=1}^T Q(g^t, g^{t-1})
\end{equation}

The total regret is the sum of the reward regret and switching regret, which is denoted by:
\begin{equation}
    \reg(T) = R_T + S_T
\end{equation}

Our goal is to develop a bandits algorithm that achieves a sublinear total regret, which implies that the round-average regret goes to 0 as $T$ goes to infinity.

The bandit problem under consideration is a combination of semi-stochastic and semi-adversarial characteristics. On the one hand, determining the optimal arm within a group presents a stochastic bandit problem. On the other hand, identifying the optimal group constitutes an adversarial bandit problem. Consequently, the problem at hand confronts uncertainties stemming from both the stochastic aspects of the environment and the adversary's adversarial behavior concurrently. The learner's received reward emerges as a combined outcome, influenced by both these factors.

\subsection{B-EXPUCB Algorithm}
In this subsection, we design a new bandit algorithm, called B-EXPUCB, to solve our considered problem, where ``B'' stands for Block. As its name suggests, B-EXPUCB marries two classical bandits algorithms, namely blocked EXP3 and LinUCB, to efficiently handle the adversarial component with switching cost and the stochastic component of the considered problem. We will prove in Section \uppercase\expandafter{\romannumeral4}-B that this marriage indeed leads to a sublinear regret bound. 

B-EXPUCB combines the strengths of both blocked EXP3 and LinUCB. On one hand, it upholds an unbiased cumulative historical reward estimate ${R}^t_g$ for each group and restricts the number of group switches by partitioning the game rounds into blocks. This division compels the algorithm to consistently select the same group for all rounds within a block, thereby streamlining group selection. Specifically, B-EXPUCB employs blocks to control $S_T$. Given a sequence of task blocks $(B_n)_{n \ge 1}$ of lengths $B$, and a time horizon $T$, we define $N$ as the smallest integer such that $\sum{n=1}^N |B_n| \ge T$. We then truncate the last block such that the cumulative length of the $N$ blocks sums up to $T$. Consequently, the blocks $B_n$ segment $T$ into $\{T_1,\dots,T_2-1\},\{T_2,\dots,T_3-1\},\dots,\{T_N,\dots,T_{N+1}-1\}$. On the other hand, B-EXPUCB maintains a linear parameter estimate $\hat{\theta}_g$ for each group, which is incrementally updated over rounds using online ridge regression. This process facilitates arm selection within a group. The algorithm's pseudo-code is outlined in Algorithm~\ref{algorithm1}, with a detailed explanation provided below.

\begin{algorithm}[t]
	\caption{B-EXPUCB}
	\begin{algorithmic}[1] 
		\State \textbf{Input}: Time horizon $T$, learning rate $\eta > 0$, and $\zeta > 0$. 
		\State \textbf{Initialization}: $V^0_{g} = \lambda I_d$, $b^0_{g} = 0$, $\hat{\theta}_g = 0$, $\forall g \in \mathcal{G}$, $n = 0$, use block lengths $B$ to segment $T$ to $\{T_1,\dots,T_2-1\},\{T_2,\dots,T_3-1\},\dots,\{T_{N+1},\dots,T\}$
		\For {$t = 1 , ..., T$}
		    \State Compute estimated cumulative reward for each $g$
		        \begin{align}
		            R_{g}^{t-1} = \sum_{s = 1}^{t-1} \frac{\textbf{1}\{g^s = g\}r^s}{P^s(g)B}
		            \label{cum-reward}
		        \end{align}
          \If{$t = T_{n+1}$}
		    \State Compute the sampling distribution for each $g$
        	    \begin{align}
        	        P^t(g) = (1-\beta)\frac{\exp(\eta R_{g}^{t-1})}{\sum_{g'=1}^G \exp(\eta  R_{g'}^{t-1})} + \frac{\beta}{G}
        	        \label{sampling}
        	    \end{align}		    
		    \State Sample group $g^t \sim P^t$, $n = n+1$
      \Else 
    \State Set $P^{t}(g) = P^{t-1}(g)$, $g^{t} = g^{t-1}$.
            \EndIf
		    \State Compute the best-estimated arm within $g^t$
		        \begin{align}
		            x^{t} = \arg\max_{x\in\mathcal{X}_{g^t}} \left((\hat{\theta}_{g^t}^{t-1})^\top x  + \alpha^t \sqrt{x^\top (V^{t-1}_{g^t})^{-1} x}\right)
		            \label{arm-estimate}
		        \end{align}		    
		    \State Play group/arm $(g^t, x^t)$
		    \State Observe reward $r^t$
		    \If {$r^t > 0$}
		        \State Update $V_{g^t}^t = V_{g^t}^{t-1} + x^t (x^{t})^\top$
		        \State Update $b_{g^t}^t = b_{g^t}^{t-1} + x^t r^t$
		        \State Update $\hat{\theta}_{g^t}^t = (V_{g^t}^t)^{-1} b_{g^t}^t$
		    \Else
		        \State $V_{g^t}^t = V_{g^t}^{t-1}$, $b_{g^t}^t = b_{g^t}^{t-1}$, $\hat{\theta}_{g^t}^t = \hat{\theta}_{g^t}^{t-1}$
		    \EndIf
		    \State For all $g \neq g^t$, $V_{g}^t = V_{g}^{t-1}$, $b_{g}^t = b_{g}^{t-1}$, $\hat{\theta}_{g}^t = \hat{\theta}_{g}^{t-1}$
		\EndFor
	\end{algorithmic}
	\label{algorithm1}
\end{algorithm}

\textbf{Group Selection}: At each round $t$, B-EXPUCB computes an unbiased estimate of the cumulative historical reward $R^{t-1}_g$ for each group $g$. This estimate is based on previous selections of groups and arms and the rewards observed, as defined in equation \eqref{cum-reward}. The indicator function $\mathbf{1}\{\cdot\}$ and the group sampling distribution from round $s$, denoted $P^s$, play crucial roles in this process. When calculating $R^{t-1}_g$, the actual reward $r^s$ received in round $s$ contributes to the cumulative reward for group $g$ only if group $g$ was the group selected in that round. This reward is then adjusted by dividing it by the probability of selecting that group. If round $t$ marks the start of a new block, it signals a transition for the learner into a fresh block, and an updated $R^{t-1}_g$ is used. At this point, a new group sampling distribution, $P^t$, is established using equation \eqref{sampling}. This distribution is formulated as a weighted sum of two components: the first, weighted by $1-\beta$, favors groups based on $\exp{(\eta R^{t-1}_g)}$, thereby prioritizing groups with higher estimated cumulative rewards (exploitation); the second, weighted by $\beta$, is a uniform distribution that treats all groups equally (exploration). The parameters $1-\beta$ and $\beta$ thus strikes the balance between exploitation and exploration at the group level. Using the defined distribution $P^t$, a group $g^t$ is sampled and designated for selection in the current round. If the learner remains within the same block, then the sampling distribution $P^t$ and the group $g^t$ remain the same with the previous rounds.

\textbf{Arm Selection}: Next, B-EXPUCB calculates the best estimated arm within the selected group $g^t$ using equation \eqref{arm-estimate}, based on the estimated group parameter $\hat{\theta}^{t-1}_g$ and the auxiliary variable $V^{t-1}_{g^t}$. The first term in \eqref{arm-estimate} quantifies the reward prediction for an arm $x$ within group $g^t$, while the second term represents the confidence level of this prediction. The parameter $\alpha^t$ plays a critical role in balancing the exploitation of known rewards and the exploration of less certain options within each group. The expression for $\alpha^t$ is defined as follows:
\begin{align}
    \alpha^t = \sqrt{\lambda} + \sigma\sqrt{d \log \frac{1 + t/\lambda}{\delta}},
\end{align}
where $\lambda$ and $\delta$ are tuning parameters of the algorithm, and $\sigma$ characterizes the noise level as described in equation \eqref{noise}.


\textbf{Variable Update}: Subsequent to playing the selected arm within the chosen group and receiving the reward, B-EXPUCB adapts its various variables based on the outcome $r^t$. Specifically, when $r^t$ is non-zero, indicating that the selected group was not subjected to an adversary attack, the auxiliary variables $V^t_{g^t}$ and $b^t_{g^t}$ are updated accordingly for the chosen group. This operation essentially constitutes an online ridge regression step, facilitating the adjustment of a new group linear parameter estimate denoted as $\hat{\theta}^t_{g^t}$. Conversely, when $r^t$ equals zero, implying that the selected group was indeed attacked by the adversary, the auxiliary variables and the linear parameter estimate remain unchanged. For the groups not selected, their auxiliary variables and linear parameter estimates likewise remain unaffected.

\subsection{Regret Analysis}
In B-EXPUCB, combating the stochastic uncertainty within a group is intertwined with combating the adversarial uncertainty across groups. Thus, the regret analysis of B-EXPUCB must consider the regrets due to these two aspects simultaneously. In this subsection, we show that through a careful selection of the algorithm parameters, B-EXPUCB achieves a sublinear regret bound.

Without loss of generality, we assume that $\|x\| \leq 1$, $\forall x \in \mathcal{X}^t_g, \forall g, t$, and $\|\theta_g\| \leq 1, \forall g$. Furthermore, we assume that the noise satisfies the $\sigma$-sub-Gaussian condition, i.e., $\forall \zeta$
\begin{align}
    \mathbb{E}[e^{\zeta n^t}|g^t, x^t, \mathcal{H}^{t-1}] \leq \exp\left(\frac{\zeta^2 \sigma^2}{2}\right).
    \label{noise}
\end{align}

We start with two lemmas on the estimation of the linear parameter of the groups. 
\begin{lemma}
The estimated reward $x^\top \hat{\theta}^t_g$ satisfies the following error bound for all arm $x$, all group $g$ and round $t$ with probability at least $1-\delta$,
\begin{align}
    |x^\top \hat{\theta}^t_g - x^\top \theta_g| \leq \alpha^t \sqrt{x^\top (V^{t}_g)^{-1}x}.
    \label{error_bound}
\end{align}
\end{lemma}
Note that Lemma 1 leads to an upper confidence bound (UCB) on the estimated reward of arms, and thus the UCB-based arm selection rule in \eqref{arm-estimate}. 

\begin{lemma}
Assume $\lambda \geq 1$, then we have for all $g$, 
\begin{align}
    &\sum_{t=1}^T \textbf{1}\{g^t = g\} a^t(g) \alpha^t \sqrt{(x^t)^\top (V^{t-1}_g)^{-1} x^t}\\
    \leq & \left(\sqrt{\lambda} + \sigma \sqrt{d\log\frac{1+T/\lambda}{\delta}}\right)\sqrt{2T d \left(\log(\lambda + \frac{T}{d}) - \log \lambda \right)}, \nonumber
\end{align}
with probability at least $1-\delta$. 
\end{lemma}

Now, we are ready to present the regret bound. 
\begin{theorem}
For any $\delta \in (0, 1)$, by choosing $\beta = T^{-1/4}\sqrt{\log(T)}$, $\eta = T^{1/6}$, and $B = T^{2/3}$, B-EXPUCB yields, with probability at least $1-\delta$, the following expected regret bounds hold:
\begin{align}
    &S_T \le O(T^{1/3}), R_T \le O(T^{3/4}\sqrt{\log(T)})
    \\&\reg(T) = O(T^{3/4}\sqrt{\log(T)}).
\end{align}
\end{theorem}
In the purely adversarial setting, where the optimal arm for each group at every round is known to the learner, the EXP3 algorithm achieves a regret bound of $O(\sqrt{T})$. Conversely, in the purely stochastic setting, where no group faces adversarial attacks in any round, LinUCB in the case of group-disjoint parameters attains a regret bound of $O(\sqrt{T\log T})$. The performance of B-EXPUCB lags behind both, due to its simultaneous handling of adversarial and stochastic uncertainties. Nevertheless, neither EXP3 nor LinUCB can secure a sublinear regret in the specific problem under consideration. When compared to EXPUCB, which sets the parameter 
$B = 1$, B-EXPUCB ensures a switching cost bound of 
$O(T^{1/3})$.


\section{SIMULATION RESULTS
}
In this section, we evaluate our proposed algorithm B-EXPUCB in the context of collaborative inference and compare its performance against several baseline
methods.

\subsection{Simulation Setup}
\begin{figure}[th]
    \centering
    \subfigure[YoLo]{
        \begin{minipage}[b]{0.49\linewidth}
        \includegraphics[width=1\textwidth]{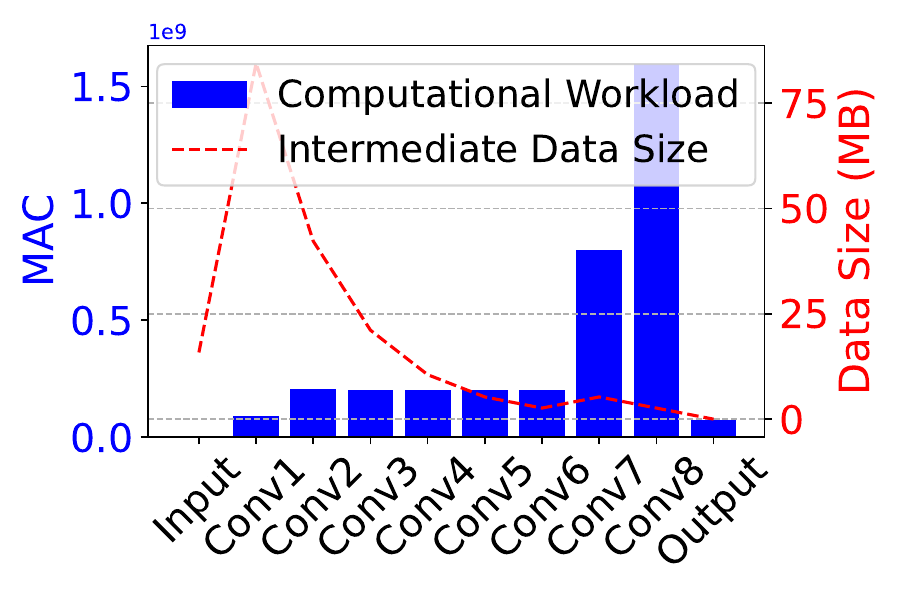}
        \end{minipage}
    \label{fig:feature of yolo}
    }
    \hspace{-6mm}
    \subfigure[ResNet50]{
        \begin{minipage}[b]{0.49\linewidth}
            \includegraphics[width=1\textwidth]{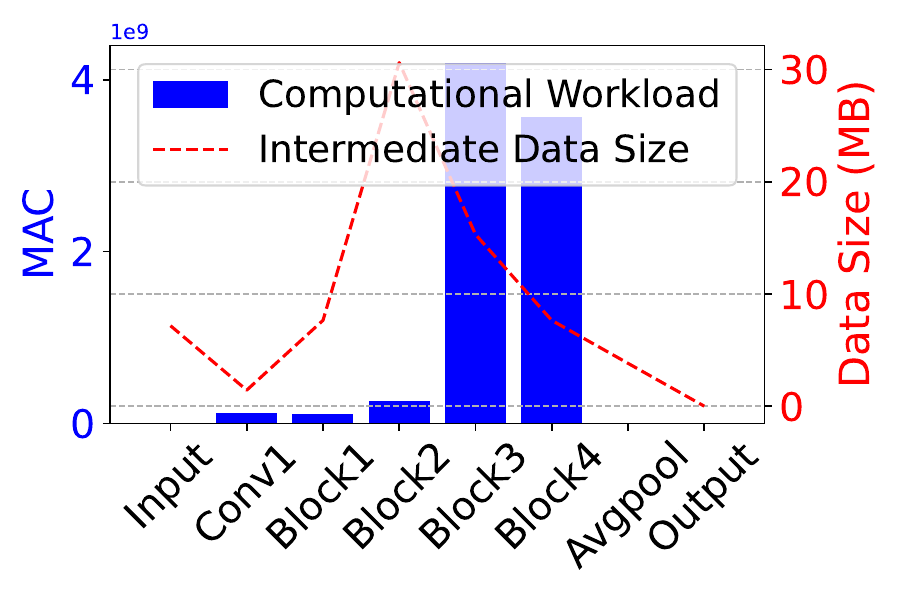}
        \end{minipage}
        \label{fig:feature of resnet}
    }\vspace{-5 pt}
    \caption{Computational workload and intermediate data size of YoLo/ResNet50.}
    \label{fig:feature of all}
    \vspace{-10 pt}
\end{figure}

\textbf{Inference Task Generation}. Each task, denoted as $t$, involves conducting inference tasks such as object detection or word prediction, using a particular DNN. In our simulations, we consider inference tasks using two widely-used DNNs, namely ResNet50\cite{he2016deep} and YoLo ~\cite{redmon2016you}. Each task $t$ is a random draw of these DNNs and hence the available arm set may change over tasks. For each DNN, we use the online analytical tool Netscope Analyzer\cite{Netscope_CNN_Analyzer} to calculate the computational workload of each layer (in terms of the number of multiply-accumulate, or MAC, units) and the intermediate data size between the layers. Thus for each potential partitioning point, the corresponding feature vector is obtained and all the possible feature vectors of a DNN are collected in a set. Fig. \ref{fig:feature of all} visualizes the computational workload and intermediate data size of YoLo/ResNet50. 
\vspace{-5 pt}

\textbf{Computing Speed and Wireless Link Rate}. In this experiment, we explore a relay network scenario where the mobile device connects to the edge server via a single relay. We model the network with 4 relays, creating $G=4$ potential paths for collaborative inference. The computing speeds of the relays and the wireless transmission rates between the mobile device and each relay are detailed in Table~\ref{server-para}, though these values remain unknown to the mobile device. We assume that the edge server does not perform computations and the connection between the relays and the server is sufficiently fast to ensure that the DNN partitioning occurs solely between the mobile device and the relay. This setup simplifies the more complex general network in our formulation, but it still allows for an effective comparison between the performance of B-EXPUCB and other baseline strategies. The mobile device's default computing speed is set at 8.255e9 MAC/s. We model the reward noise with a uniform distribution over the range $[-0.05, 0.05]$.


\begin{table}
\centering
\caption{Relay Path Parameters}
\begin{tabular}{|c|c|c|c|c|} 
\hline
        & Relay 1 & Relay 2 & Relay 3 & Relay 4  \\ 
\hline
Comp.
Speed (MAC/s) & 4.125e10      & 4.125e10      & 8.25e10       & 8.25e10\\ 
\hline
Link
Rate (Mbps)         & 50            & 40            & 20            & 10\\
\hline
\end{tabular}
\label{server-para}
\vspace{-10pt}
\end{table}


\textbf{Adversary's Strategy}. We simulate both an oblivious attacking strategy and an adaptive attacking strategy. For the oblivious strategy, the adversary uses a randomized Markov attacking strategy, whose transition matrix is
\begin{equation}
\begin{bmatrix}
0.4 & 0.1 & 0.4 & 0.1 \\
0.35 & 0.15 & 0.35 & 0.15\\
0.4 & 0.1 & 0.4 & 0.1\\
0.35 & 0.15 & 0.35 & 0.15
\end{bmatrix},\nonumber
\end{equation}
where the element in row $i$ and column $j$ represents the probability of attacking path $j$ at task $t$ if path $i$ was attacked at task $t-1$. For the adaptive strategy,  the adversary observes the mobile device's path selection decision at task $t-1$, and attacks the same path at task $t$. In both strategies, exactly one path is attacked for each task. 

\textbf{Switching Cost.} In this setting, We incorporate the switching cost function as $Q(g^t, g^{t-1}) = \textbf{1}\{g^t \neq g^{t-1}\}$ by setting $\tau = 1$. 

\subsection{Baseline Algorithms}

In addition to the \textbf{Oracle} strategy described in \eqref{optimal}, we examine the following four baseline approaches:

\textbf{EXPUCB (No Switching Cost)}: This variant of B-EXPUCB, as introduced in our previous work \cite{huang2023adversarial}, sets the hyperparameter $B$ to 1, disregarding switching costs. Consequently, EXPUCB allows for group changes in every time slot, without imposing any block structure.

\textbf{LinUCB (Group-Based)}: This approach applies the classical LinUCB algorithm in the context of group selection, without considering attacks. In each task, it selects the group and arm with the highest UCB estimate for reward, as defined in \eqref{arm-estimate}. Auxiliary variables remain unchanged when the received reward is 0.

\textbf{EXP3 (Fixed Arm Assignment)}: This utilizes the standard EXP3 algorithm for group selection. Each group is associated with a fixed arm throughout all tasks. In our simulations, we assign a default arm that leverages the chosen relay to process all computation workload.

\textbf{Local}: This strategy involves retaining all computation tasks on the mobile device itself, eliminating the need for learning or exposure to attacks. 

For B-EXPUCB, the default algorithm parameters are $\delta = 0.1$, $\lambda = 1$,$d=3$, $\sigma = 0.05$, $\beta = T^{-1/4}\sqrt{\log(T)}$, $\eta = T^{1/6}$, $B = T^{2/3}$ , and $\alpha = \sqrt{\lambda} + \sigma \sqrt{d\log\frac{1+T/\lambda}{\delta}}$. 

\subsection{Performance Comparison}

\begin{figure*}[th]
	\centering
	\subfigure[Only YoLo Tasks]{
		\begin{minipage}[b]{0.3\textwidth}
			\includegraphics[width=0.95\textwidth]{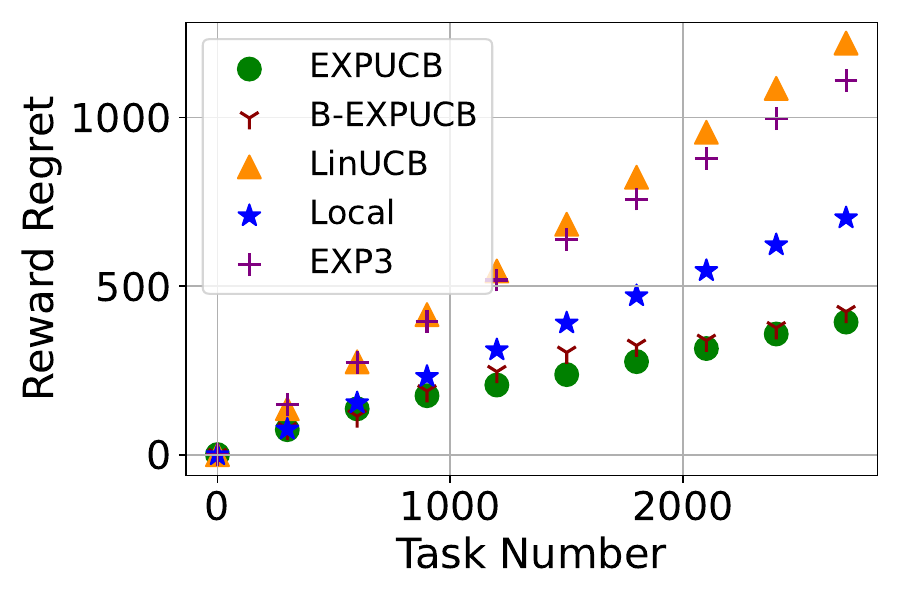}
		\end{minipage}
		\label{fig:regretYoLo}
	}
	\hspace{-6mm}
    	\subfigure[Only ResNet50 Tasks]{
    		\begin{minipage}[b]{0.3\textwidth}
   		 	\includegraphics[width=0.95\textwidth]{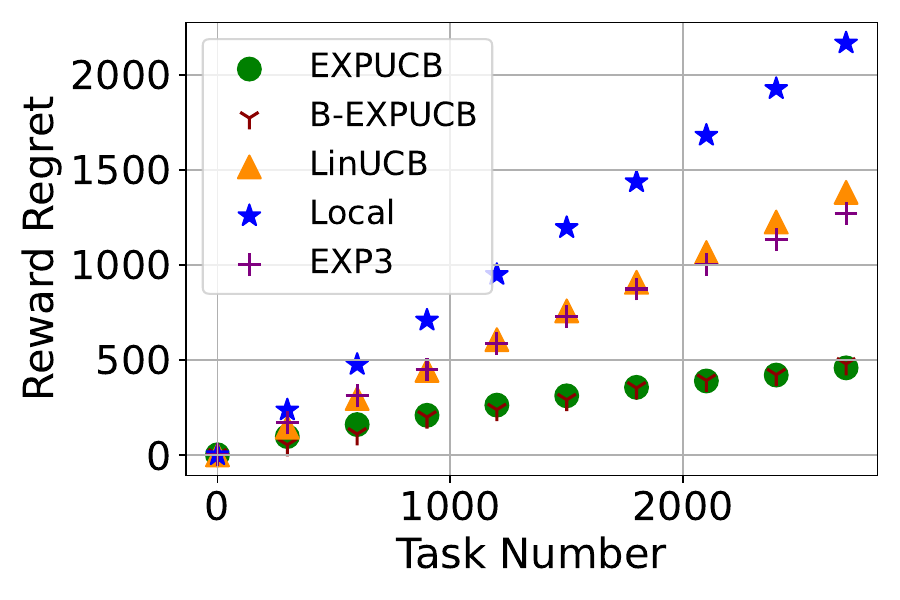}
    		\end{minipage}
		\label{fig:regretResNet}
    	}
    \hspace{-6mm}
    	\subfigure[Half YoLo Half ResNet50 Tasks]{
		    \begin{minipage}[b]{0.3\textwidth}
   	 	    \includegraphics[width=0.95\textwidth]{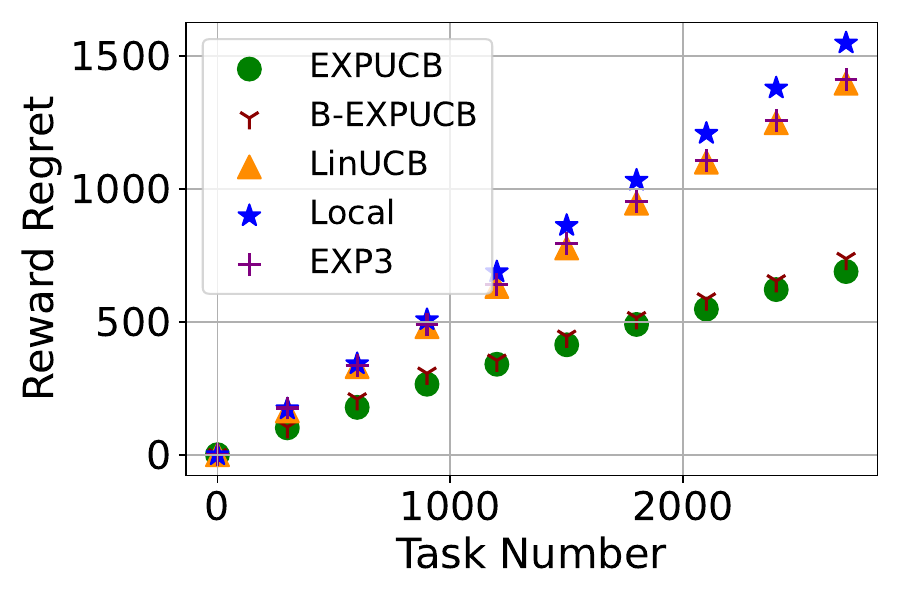}
		    \end{minipage}
	    \label{fig:regretMix}
	    }
	\caption{Reward Regret achieved by B-EXPUCB and baselines.}
	\label{fig:regret}
    \vspace{-10pt}
\end{figure*}

\begin{figure*}[th]
	\centering
	\subfigure[Only YoLo Tasks]{
		\begin{minipage}[b]{0.3\textwidth}
			\includegraphics[width=0.95\textwidth]{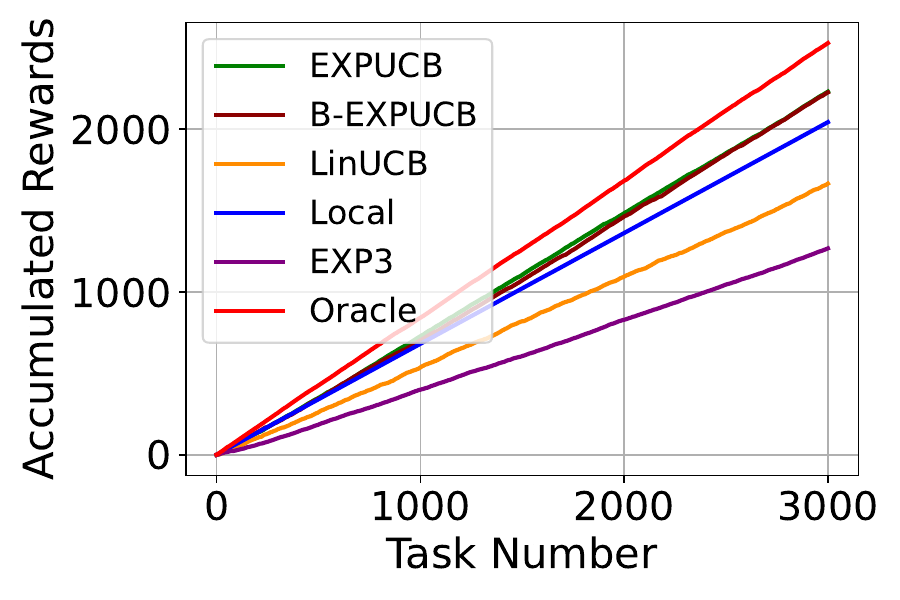}
		\end{minipage}
		\label{fig:rewardYoLo}
	}
	\hspace{-6mm}
    	\subfigure[Only ResNet50 Tasks]{
    		\begin{minipage}[b]{0.3\textwidth}
   		 	\includegraphics[width=0.95\textwidth]{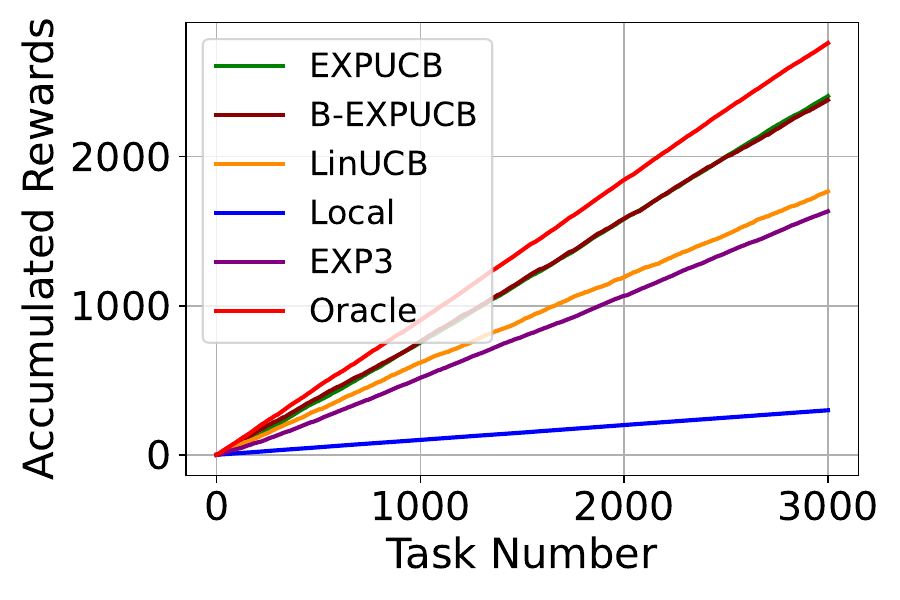}
    		\end{minipage}
		\label{fig:rewardResNet}
    	}
    \hspace{-6mm}
    	\subfigure[Half YoLo Half ResNet50 Tasks]{
		    \begin{minipage}[b]{0.3\textwidth}
   	 	    \includegraphics[width=0.95\textwidth]{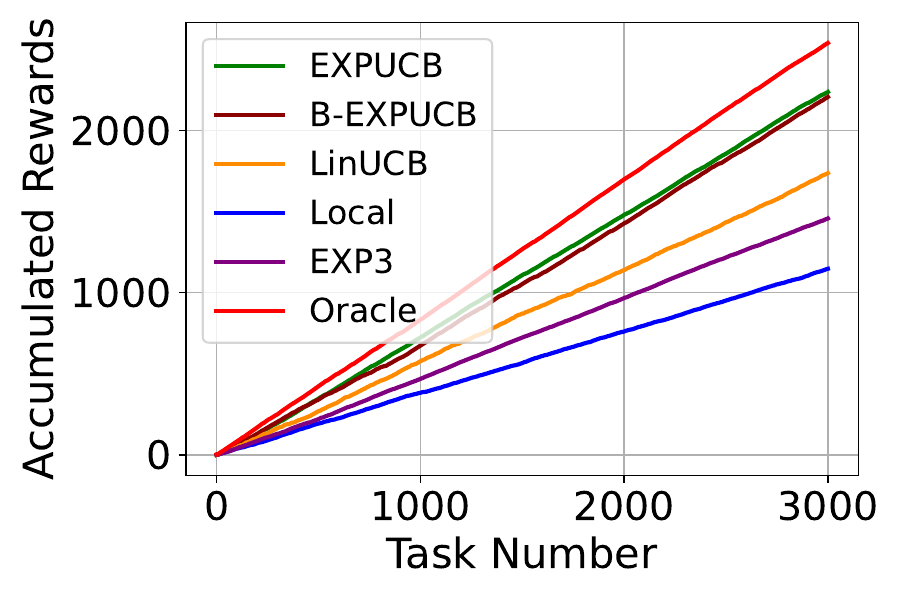}
		    \end{minipage}
	    \label{fig:rewardMix}
	    }
	\caption{Total reward achieved by B-EXPUCB and baselines.}
	\label{fig:reward}
    \vspace{-10pt}
\end{figure*}

\begin{figure*}[th]
	\centering
	\subfigure[Only YoLo Tasks]{
		\begin{minipage}[b]{0.3\textwidth}
			\includegraphics[width=0.95\textwidth]{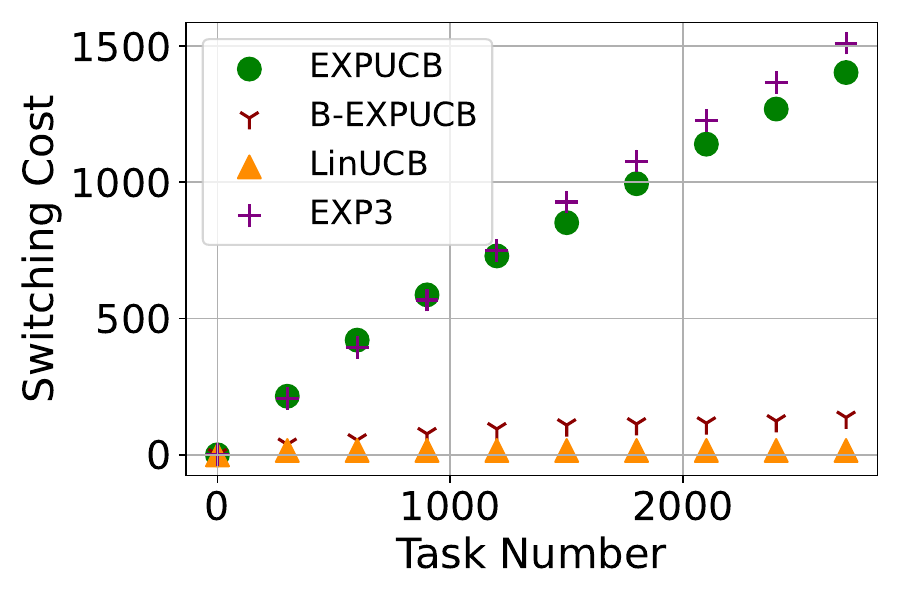}
		\end{minipage}
		\label{fig:switchingcostYoLo}
	}
	\hspace{-6mm}
    	\subfigure[Only ResNet50 Tasks]{
    		\begin{minipage}[b]{0.3\textwidth}
   		 	\includegraphics[width=0.95\textwidth]{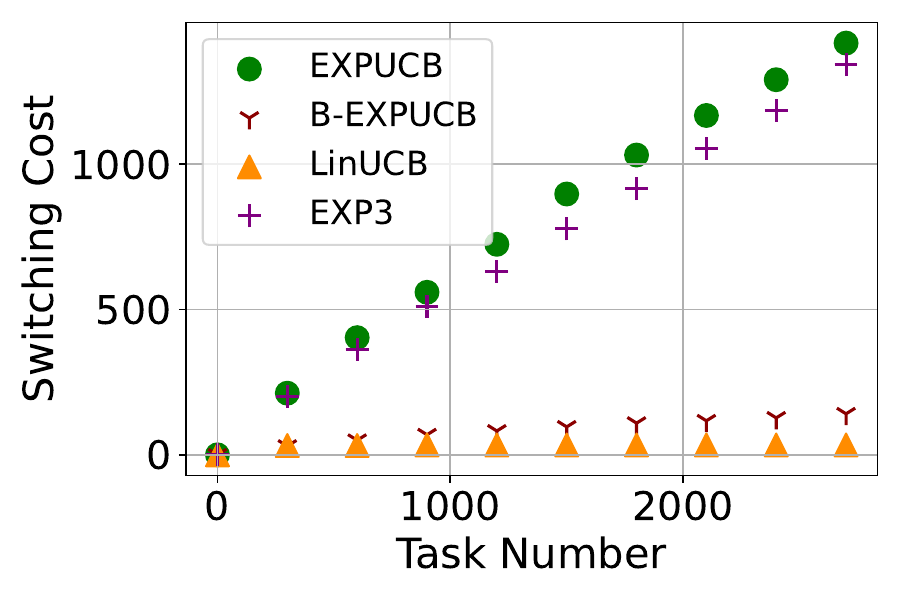}
    		\end{minipage}
		\label{fig:switchingcostResNet}
    	}
    \hspace{-6mm}
    	\subfigure[Half YoLo Half ResNet50 Tasks]{
		    \begin{minipage}[b]{0.3\textwidth}
   	 	    \includegraphics[width=0.95\textwidth]{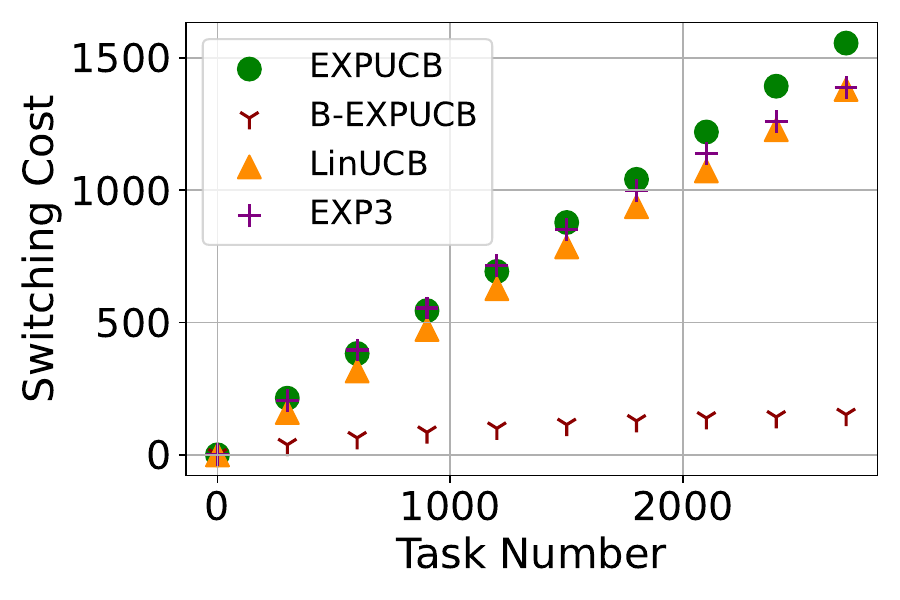}
		    \end{minipage}
	    \label{fig:switchingcostMix}
	    }
	\caption{Total switching cost achieved by B-EXPUCB and baselines.}
	\label{fig:switchingcost}
    \vspace{-10pt}
\end{figure*}

We first compare the performance of B-EXPUCB with the baselines in terms of the total reward regret (shown in Fig.  \ref{fig:regret}), the total reward (shown in Fig. \ref{fig:reward}) and total switching cost (shown in Fig.  \ref{fig:switchingcost}) under the oblivious attacker. Three task distributions are simulated: all tasks are YoLo; all tasks are ResNet50; a task is YoLo with probability 0.5 and ResNet50 with probability 0.5. Note that Oracle may change for different numbers of tasks and hence, we present only a few points in Fig. \ref{fig:regret}. The Oracle presented in Fig. \ref{fig:reward} is with respect to $T = 3000$ only. In all scenarios, B-EXPUCB and EXPUCB outperform the other non-Oracle baselines and exhibit a sublinear regret in the number of tasks. The reason why the non-Oracle baselines underperform EXPUCB and B-EXPUCB is as follows. \textbf{Local}: Although Local avoids any attacks by processing all computation workload on the mobile device, it misses the benefit of leveraging a more powerful relay. Particularly, when the task is ResNet50, which is a more complex DNN model than YoLo, \textbf{Local} results in a much worse performance than the other algorithms as shown in Fig. \ref{fig:rewardResNet}. \textbf{EXP3}: EXP3 utilizes the more powerful relay by offloading all workload of a task to the selected relay. However, even though EXP3 can adapt to the attacking strategy, it fails to find the optimal workload distribution among the mobile device and the relays. \textbf{LinUCB}: LinUCB works the best among the non-Oracle baselines thanks to its ability to learn the optimal DNN partition for each relay. However, since it neglects the attacks, the path selected by LinUCB may experience many attacks, resulting in lower accumulative rewards than EXPUCB and B-EXPUCB. The accumulative rewards of B-EXPUCB are slightly worse than those of EXPUCB but show a much lower accumulated switching cost than that of EXPUCB. This is because the B-EXPUCB uses the block size and is forced to select the same path in the block, which decreases the switching cost. Besides, although EXPUCB can change the path at each task round, it also explores the other non-optimal paths. Once B-EXPUCB selects the optimal group and does not change it in the block, it offsets the produced regret when always selecting the non-optimal path in the other blocks. 

\subsection{Behaviors of B-EXPUCB}
\begin{figure*}[th]
	\centering
 	\begin{minipage}[b]{0.72\textwidth}
	\hspace{-4mm}
	\subfigure[Only YoLo Tasks]{
		\begin{minipage}[b]{0.33\textwidth}
			\includegraphics[width=1\textwidth]{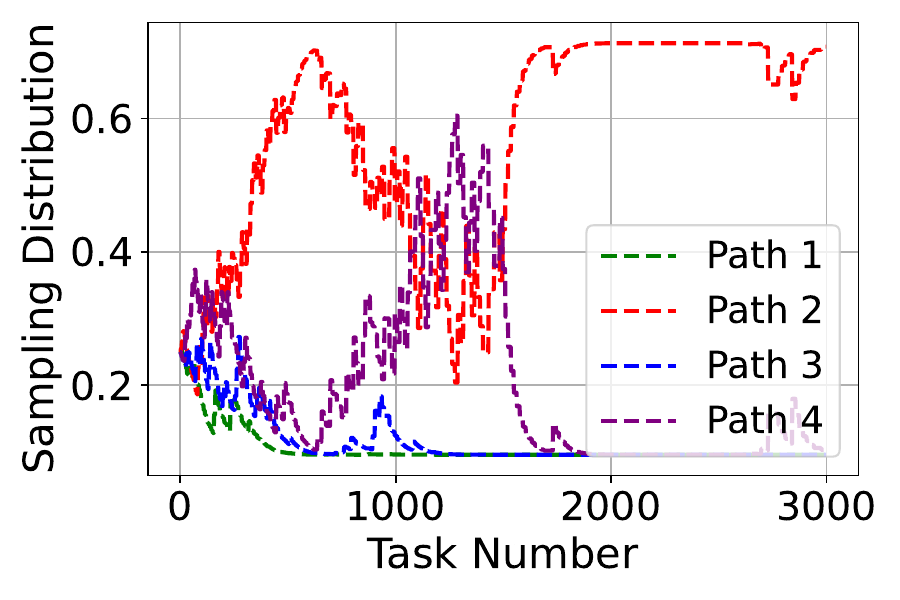}
		\end{minipage}
		\label{fig:samplingProbYoLo}
	}
	\hspace{-4mm}
    	\subfigure[Only ResNet50 Tasks]{
    		\begin{minipage}[b]{0.33\textwidth}
   		 	\includegraphics[width=1\textwidth]{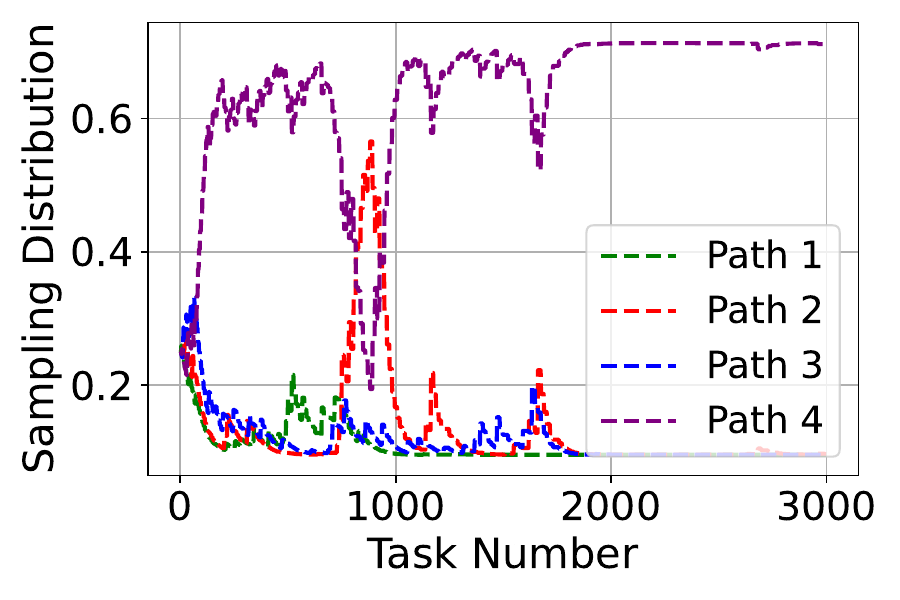}
    		\end{minipage}
		\label{fig:samplingProbResNet}
    	}
    \hspace{-4mm}
    	\subfigure[Half YoLo Half ResNet50 Tasks]{
		    \begin{minipage}[b]{0.33\textwidth}
   	 	    \includegraphics[width=1\textwidth]{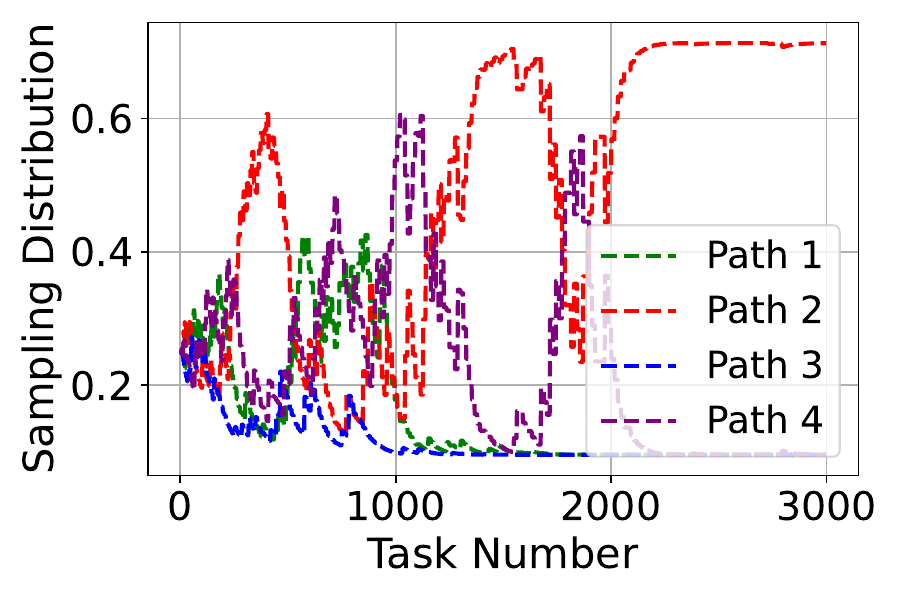}
		    \end{minipage}
	    \label{fig:samplingProbMix}
	    }
	\caption{Evolution of the sampling distribution of B-EXPUCB for each Path.}
	\label{samplingProb}
	 \vspace{-10 pt}
	\end{minipage}
	\begin{minipage}[b]{0.25\textwidth}
	\hspace{-4mm}
   	 	\includegraphics[width=1\textwidth]{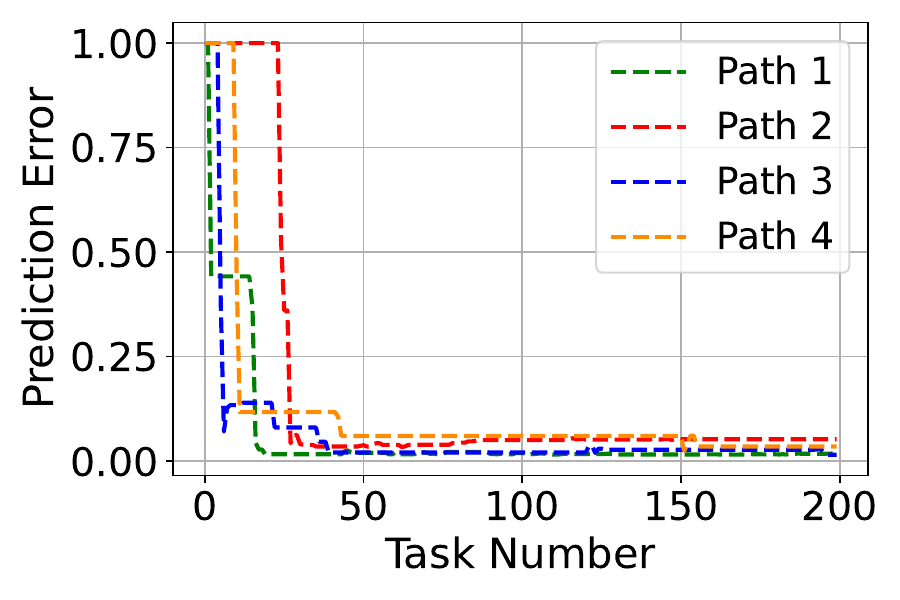}
   	\caption{Parameter prediction error \\(Half YoLo Half ResNet50 Tasks).}
   	\label{fig:predictionErrorMix}
	\end{minipage}
 \vspace{-10 pt}
\end{figure*}

Next, we delve into a more detailed analysis of the behaviors exhibited by EXPUCB and B-EXPUCB. First, Figure \ref{fig:predictionErrorMix} illustrates the progression of prediction errors associated with the group/path parameters of B-EXPUCB as the system processes an increasing number of tasks. Specifically, this figure focuses on the scenario where the tasks are an equal mix of half-YoLo and half-ResNet50 tasks. It is notable that even when an attacker attempts to compromise the task processing by targeting the paths, thereby reducing the opportunities for the learner to glean insights into the path parameters, B-EXPUCB demonstrates remarkable agility in learning these parameters rapidly. This translates into a low prediction error that is achieved within the initial 100 tasks. Although similar trends are observed for YoLo-only and ResNet50-only tasks, we omit those specific results here for the sake of brevity and space conservation.

\begin{figure*}[th]
	\centering
	\subfigure[Only YoLo Tasks]{
		\begin{minipage}[b]{0.3\textwidth}
			\includegraphics[width=0.95\textwidth]{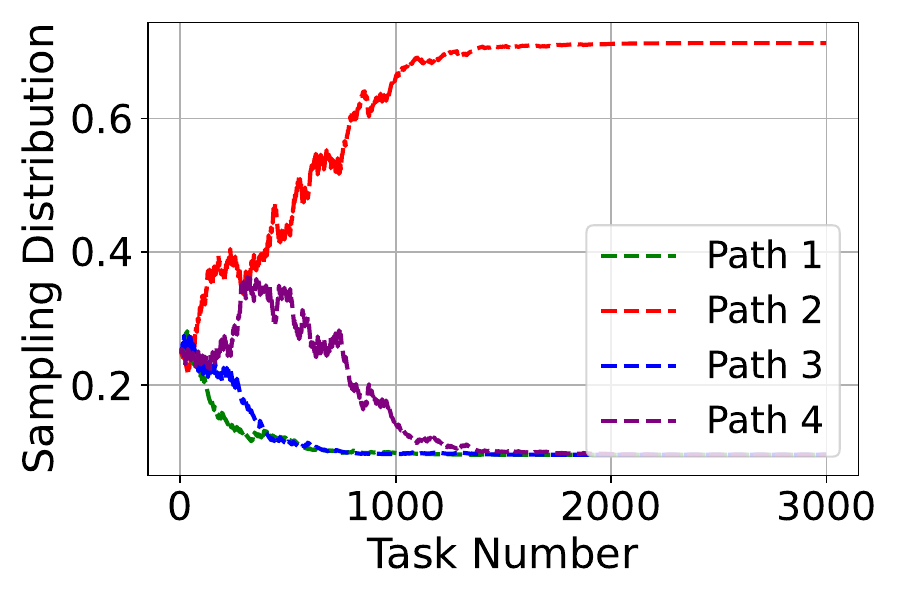}
		\end{minipage}
		\label{fig:samplingProbYoLo_expucb}
	}
	\hspace{-6mm}
    	\subfigure[Only ResNet50 Tasks]{
    		\begin{minipage}[b]{0.3\textwidth}
   		 	\includegraphics[width=0.95\textwidth]{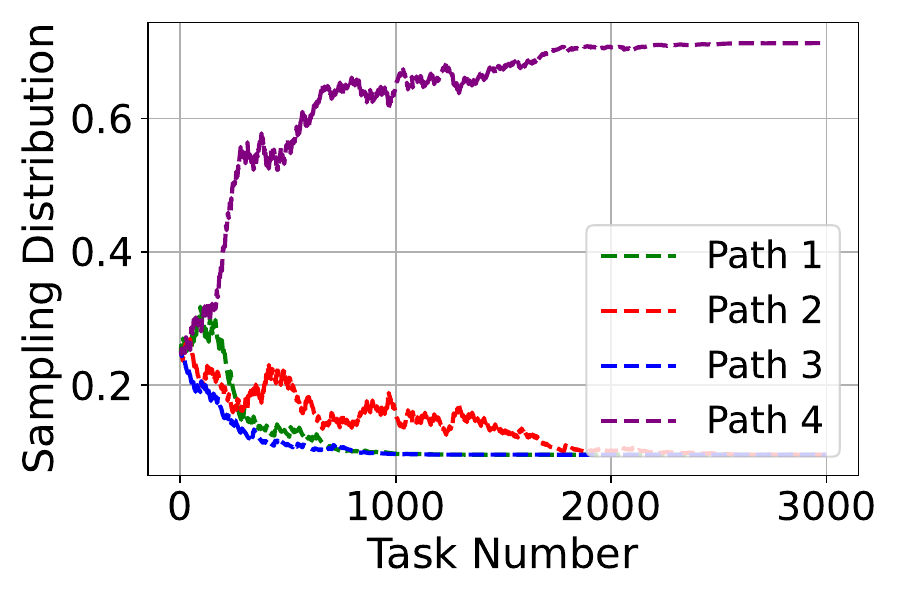}
    		\end{minipage}
		\label{fig:samplingProbResNet_expucb}
    	}
    \hspace{-6mm}
    	\subfigure[Half YoLo Half ResNet50 Tasks]{
		    \begin{minipage}[b]{0.3\textwidth}
   	 	    \includegraphics[width=0.95\textwidth]{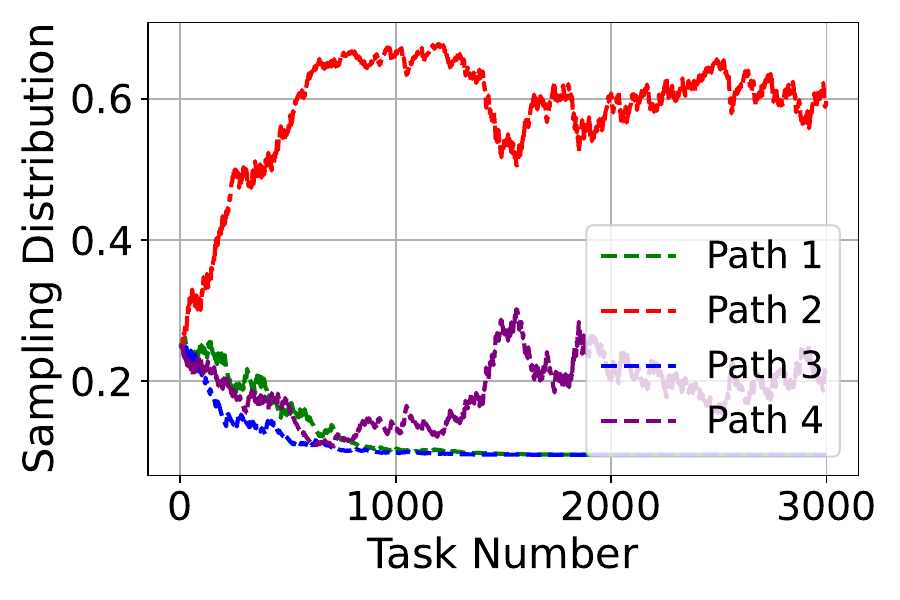}
		    \end{minipage}
	    \label{fig:samplingProbMix_expucb}
	    }
	\caption{Evolution of the sampling distribution of EXPUCB for each Path.}
	\label{samplingProb_expucb}
    \vspace{-10pt}
\end{figure*}

Next, Fig. \ref{samplingProb_expucb} and \ref{samplingProb} show the evolution of the group/path sampling distributions of EXPUCB and B-EXPUCB separately. According to the characteristics of YoLo and ResNet50 tasks, our simulation was designed in a way so that Paths 1 and 2 are the preferred paths for YoLo tasks (Path 1 being the best), and Paths 3 and 4 are the preferred paths for ResNet50 tasks (Paht 3 being the best) in the attack-free scenario. As Fig. \ref{fig:samplingProbYoLo_expucb} and \ref{fig:samplingProbYoLo} show, EXPUCB and B-EXPUCB successfully identify Path 2 as a preferred path for YoLo-only tasks. Moreover, EXPUCB and B-EXPUCB do not choose Path 1 with a high probability, despite that Path 1 is the best in the attack-free scenario because it also successfully detects that Path 1 suffers more attacks. Likewise, Fig. \ref{fig:samplingProbResNet_expucb} and \ref{fig:samplingProbResNet} shows that EXPUCB and B-EXPUCB choose Path 4 with the highest probability for ResNet50-only tasks because Path 3, which is the best in the attack-free scenario, suffers more attacks. For the half-YoLo half-ResNet50 tasks case, Paths 1 and 2 achieve higher overall rewards than Paths 3 and 4 without attacks. As Fig. \ref{fig:samplingProbMix_expucb} and \ref{fig:samplingProbMix} shows, EXPUCB and B-EXPUCB identify Path 2 as the best choice under attacks, which is the same as the optimal path in Oracle during 3000 tasks. 

Another observation is that the sampling probability of selecting the optimal path converges faster than that of B-EXPUCB, this is because the EXPUCB can change the selected path at each timeslot but B-EXPUCB only fixes the same path in the block, therefore, B-EXPUCB needs more tasks to learn the attack probability of each path.


\subsection{Impact of Time-Varying Task Distributions}
In this set of experiments, we test B-EXPUCB in scenarios where the task distributions may change over time. In the first 2000 tasks, a task is YoLo with probability 0.8 and ResNet50 with probability 0.2. From task 2000 on, a task is YoLo with probability 0.2 and ResNet50 with probability 0.8. Therefore, there is a sudden change in the task distribution at task 2000. 

\begin{figure}[th]
    \centering
    \subfigure[Accumulated rewards]{
        \begin{minipage}[b]{0.49\linewidth}
        \includegraphics[width=0.95\textwidth]{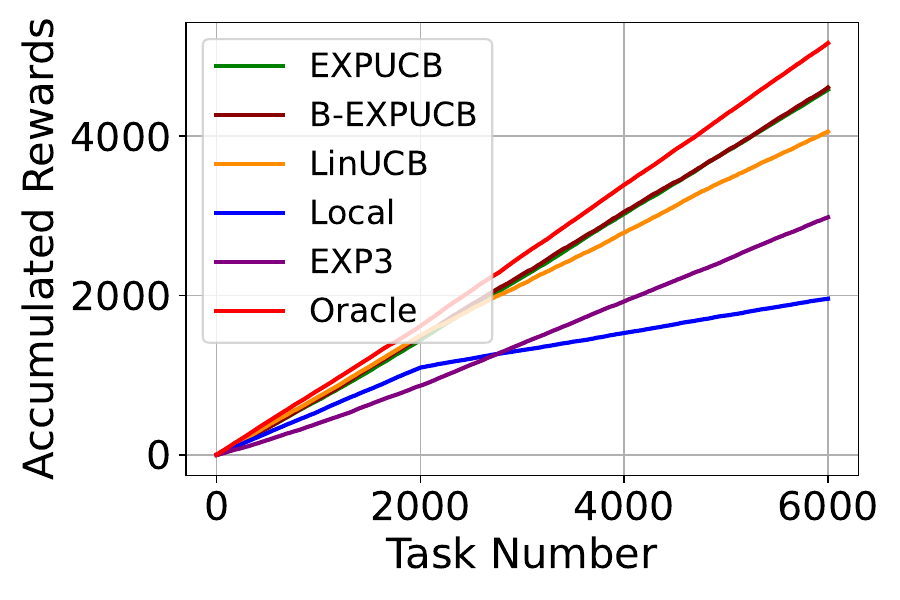}
        \end{minipage}
    \label{fig:rewardDynamicTasks}
    }
    \hspace{-6mm}
    \subfigure[Accumulated switching cost]{
        \begin{minipage}[b]{0.49\linewidth}            \includegraphics[width=0.95\textwidth]{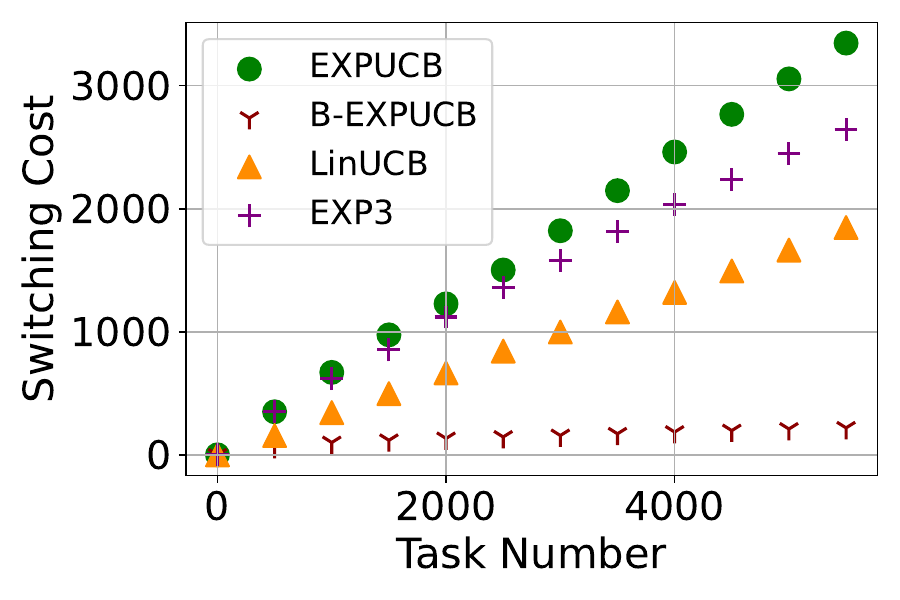}
        \end{minipage}
        \label{fig:samplingProbDynamicTasks}
    }
    \caption{Performance of B-EXPUCB under changing task distributions.}
    \label{fig:dynamicArrivalTasks}
    \vspace{-10 pt}
\end{figure}

\begin{figure}[th]
    \centering
        \subfigure[Sampling dis. of EXPUCB]{
        \begin{minipage}[b]{0.49\linewidth}
            \includegraphics[width=0.95\textwidth]{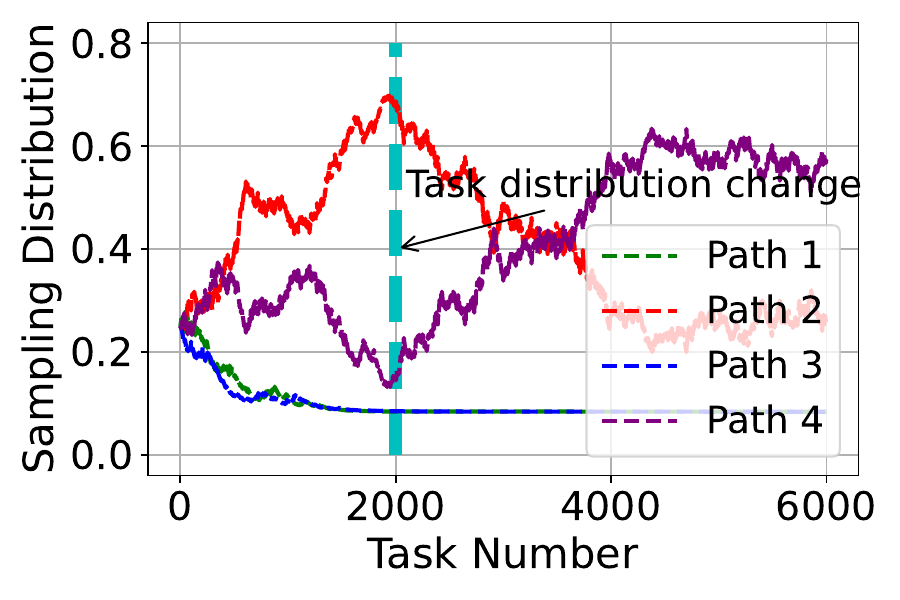}
        \end{minipage}
        \label{fig:samplingProbDynamicArrivalTasks_expucb}
    }
    \hspace{-6mm}
    \subfigure[Sampling dis. of B-EXPUCB]{
        \begin{minipage}[b]{0.49\linewidth}
        \includegraphics[width=0.95\textwidth]{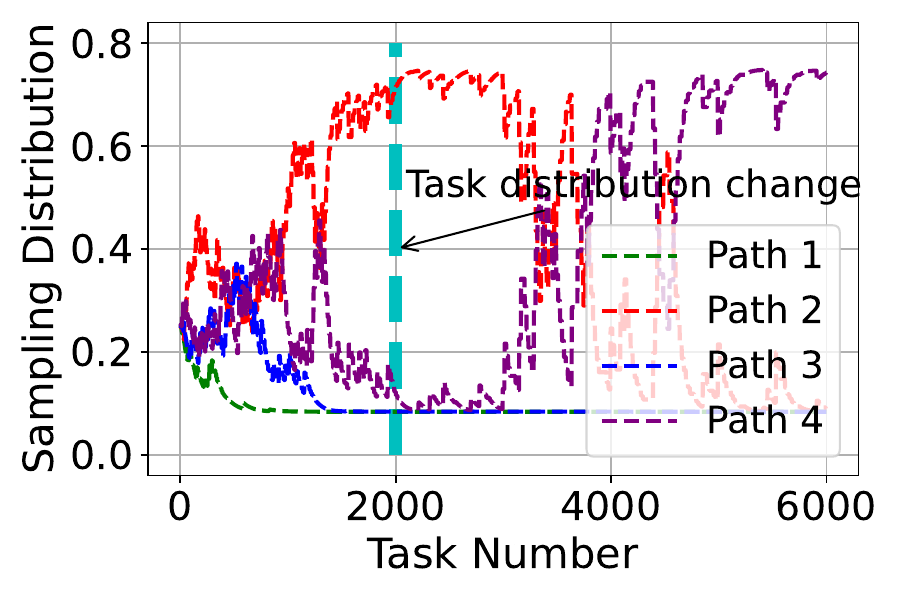}
        \end{minipage}
    \label{fig:samplingProbDynamicArrivalTasks_bexpucb}
    }\vspace{-5 pt}
    \caption{Evolution of the sampling distribution of EXPUCB VS B-EXPUCB for each path under the changing task distribution.}
    \label{fig:samplingProb_dynamicArrivalTasks}
    \vspace{-15 pt}
\end{figure}

Fig. \ref{fig:dynamicArrivalTasks} shows the performance of B-EXPUCB in this time-varying task distribution setting. As shown in Fig. \ref{fig:dynamicArrivalTasks}(a), EXPUCB and B-EXPUCB still achieve the highest total reward compared to the other non-Oracle baselines since they are able to learn the change and make adaptive decisions. Fig. \ref{fig:dynamicArrivalTasks}(b) shows that the B-EXPUCB shows much lower switching costs than that of EXPUCB. This is supported by Fig. \ref{fig:samplingProb_dynamicArrivalTasks}, which demonstrates that their path sampling distributions gradually change after the task distribution change occurs. In particular, in the first 2000 tasks, EXPUCB and B-EXPUCB select path 2 with the highest probability but after task 2000, path 4 emerges as the preferred choice of EXPUCB and B-EXPUCB. Besides, the path sampling distribution of B-EXPUCB changes more slowly than that of EXPUCB because the B-EXPUCB fixes the same path in each block.

\vspace{-5 pt}

\subsection{Impact of Attacker Strategies}

\begin{figure}[th]
	\centering  
	\subfigure[Accumulated rewards]{
		\label{fig:rewardDynamicMarkov}
        \begin{minipage}[b]{0.49\linewidth}		
		\includegraphics[width=0.95\linewidth]{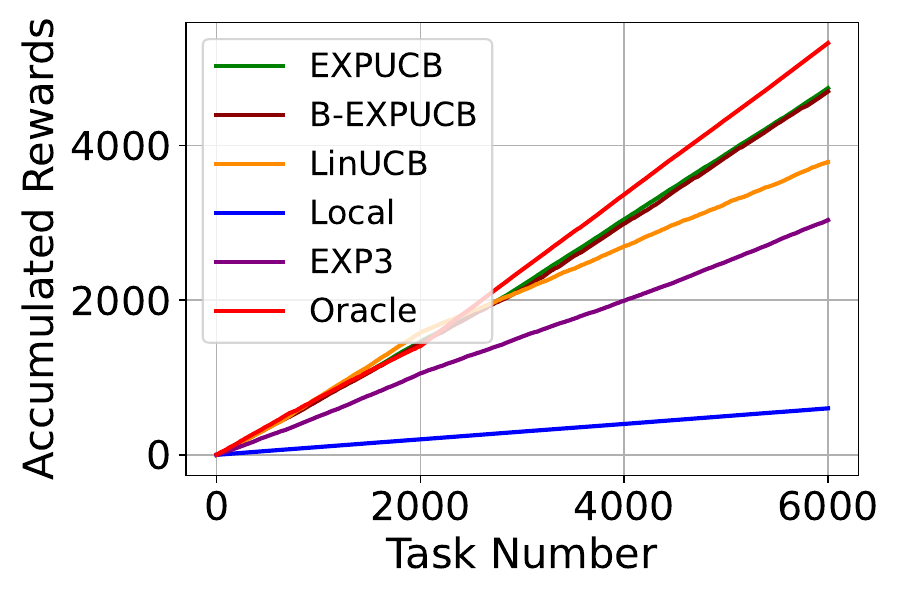}
		\end{minipage}
		}
	\hspace{-6mm}
	\subfigure[Accumulated switching cost]{
		\label{fig:switchingDynamicMarkov}		        \begin{minipage}[b]{0.49\linewidth}
		\includegraphics[width=0.95\linewidth]{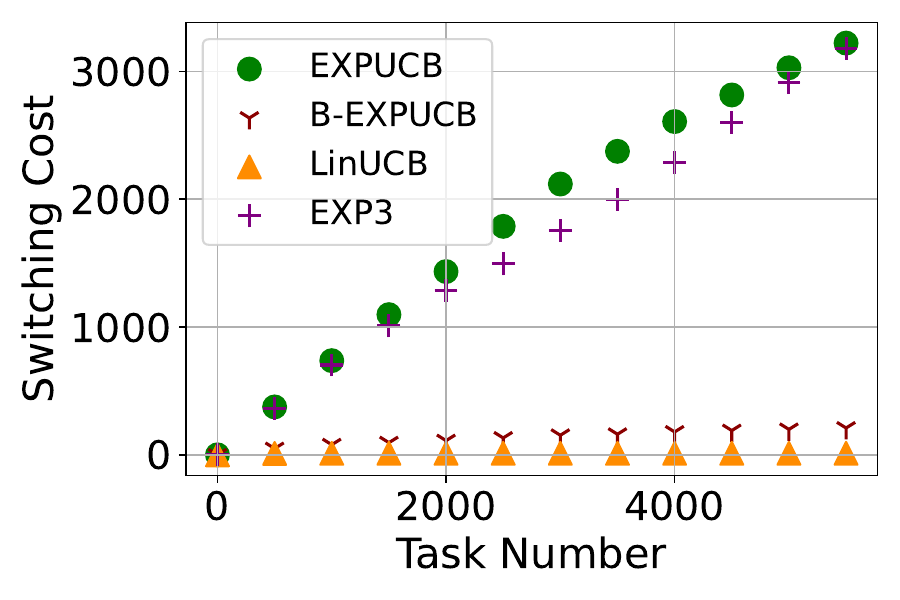}
		\end{minipage}
		}
	\caption{Performance of B-EXPUCB under the dynamic Markov attacking strategy. (ResNet50-only tasks).}
	\label{fig:dynamicMarkov}
\end{figure}

\begin{figure}[th]
    \centering
        \subfigure[Sampling dis. of EXPUCB]{
        \begin{minipage}[b]{0.49\linewidth}
            \includegraphics[width=0.95\textwidth]{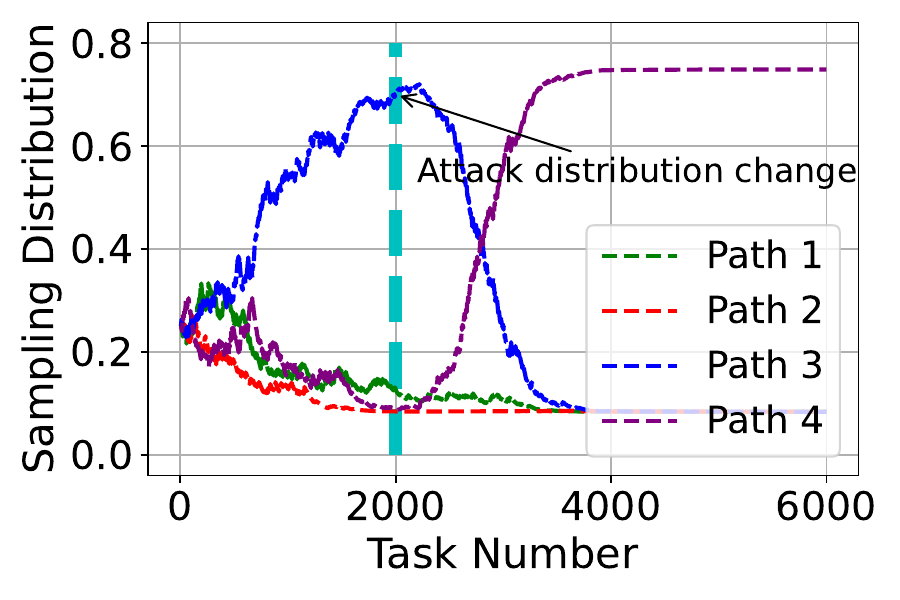}
        \end{minipage}
        \label{fig:samplingProbDynamicMarkov_expucb}
    }
    \hspace{-6mm}
    \subfigure[Sampling dis. of B-EXPUCB]{
        \begin{minipage}[b]{0.49\linewidth}
        \includegraphics[width=0.95\textwidth]{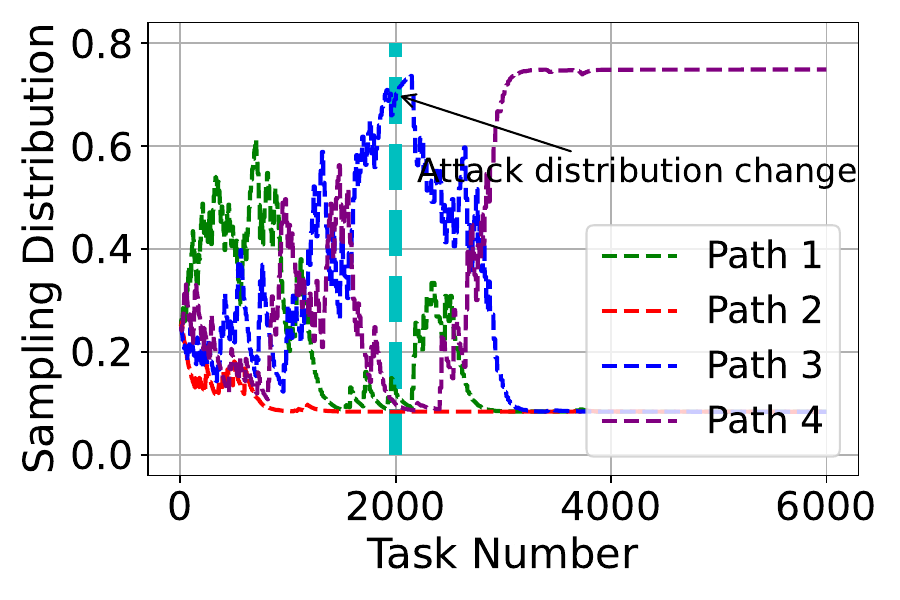}
        \end{minipage}
    \label{fig:samplingProbDynamicMarkov_bexpucb}
    }
    \caption{Evolution of the sampling distribution of EXPUCB VS B-EXPUCB for each path under the dynamic Markov attacking strategy (ResNet50-only tasks).}
    \label{fig:samplingProbDynamicMarkov}
\end{figure}

\textbf{Time-varying Attacking Strategy}. In the previous experiments, we studied the performance of B-EXPUCB under the oblivious attacking strategy. In this subsection, we consider a time-varying attacking strategy. We simulated a scenario where an oblivious attacker performs time-varying strategies to understand how B-EXPUCB adapts to the attacking strategy change. Specifically, the attacker changes its attacking transition matrix at task 2000 as follows:
\begin{equation}
\begin{bmatrix}
0.2 & 0.3 & 0.2 & 0.3 \\
0.15 & 0.35 & 0.15 & 0.35\\
0.2 & 0.3 & 0.2 & 0.3\\
0.15 & 0.35 & 0.15 & 0.35
\end{bmatrix} \Rightarrow
\begin{bmatrix}
0.35 & 0.15 & 0.45 & 0.05 \\
0.3 & 0.2 & 0.4 & 0.1\\
0.35 & 0.15 & 0.45 & 0.05\\
0.3 & 0.2 & 0.4 & 0.1
\end{bmatrix}.\nonumber
\end{equation}
Fig. \ref{fig:rewardDynamicMarkov} shows the total reward received by different algorithms, where Oracle is computed with respect to the total 6000 tasks. In the first 2000 tasks, LinUCB actually outperforms EXPUCB and B-EXPUCB because the best path in the attack-free scenario (i.e. Path 3) coincides with the path under the least attack in the simulation. Therefore, LinUCB selects the best path for more tasks than EXPUCB and B-EXPUCB. However, after the attacking strategy changes, Path 3 is no longer the best path but LinUCB fails to make the adaptation. Therefore, EXPUCB and B-EXPUCB outperform LinUCB overall. Fig. \ref{fig:switchingDynamicMarkov} shows that BEXPUCB achieves lower accumulative switching costs than that of EXPUCB. Fig. \ref{fig:samplingProbDynamicMarkov} shows that the path selection probability clearly changes after task 2000, again confirming the adaptation ability of EXPUCB and B-EXPUCB. 

\textbf{Adaptive Attacking Strategy}. We also study B-EXPUCB under the adaptive attacking strategy where the adversary attacks the path selected by the learner for the last task. Note that LinUCB in this adaptive attacking scenario would perform extremely badly because it is not adaptive to attacks and can easily be trapped in a path that receives constant attacks according to the adaptive attacking strategy. Therefore we also consider a stronger version of LinUCB where the learner randomly selects a path for each task and then selects the partition according to LinUCB. We label this additional baseline LinUCB-Random. 

\begin{figure}[th]
    \centering
        \subfigure[Sampling dis. of EXPUCB]{
        \begin{minipage}[b]{0.49\linewidth}
            \includegraphics[width=0.95\textwidth]{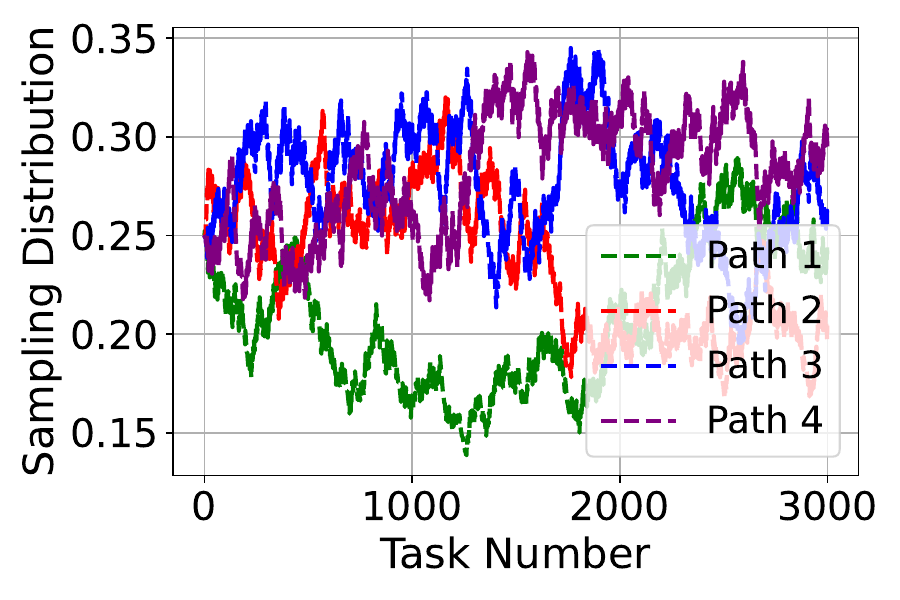}
        \end{minipage}
        \label{fig:samplingProbAdaptiveAttacker_expucb}
    }
    \hspace{-6mm}
    \subfigure[Sampling dis. of B-EXPUCB]{
        \begin{minipage}[b]{0.49\linewidth}
        \includegraphics[width=0.95\textwidth]{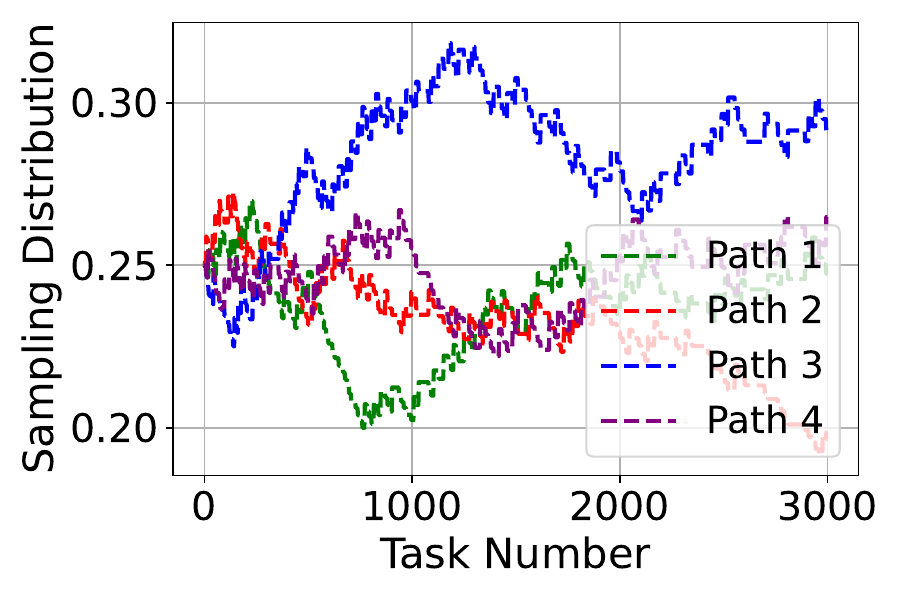}
        \end{minipage}
    \label{fig:samplingProbAdaptiveAttacker_bexpucb}
    }\vspace{-5 pt}
    \caption{Evolution of the sampling distribution of EXPUCB VS B-EXPUCB for each path under the adaptive attacking strategy (ResNet50-only tasks).}
    \label{fig:samplingProbAdaptiveAttacker}
    \vspace{-10 pt}
\end{figure}

Fig. \ref{fig:samplingProbAdaptiveAttacker} shows the total reward and switching cost obtained by B-EXPUCB and the baselines in the setting where all tasks are ResNet50. As can be seen, B-EXPUCB exhibits low accumulative switching costs at the expense of the losing reward on the adaptive attacker and the performance gap of EXPUCB is larger in this adaptive attacker case than in the oblivious attacker case. This demonstrates the superior abilities of EXPUCB to adapt the path and partition selection decisions at the trade-off between the accumulative rewards and accumulative switching costs in more challenging scenarios. Fig. \ref{fig:samplingProbAdaptiveAttacker} further illustrates the path sampling probabilities of EXPUCB and B-EXPUCB over time. As we mentioned before, in the ResNet50-only task setting, Paths 3 and 4 are the preferred paths. This is consistent with our simulation results where Paths 3 and 4 are selected with higher probabilities. Notably, EXPUCB and B-EXPUCB randomize between these two paths instead of sticking to the best one found in the attack-free scenario (i.e., Path 3) in order to escape from the constant attacks resulting from the adaptive attacking strategy. In addition, the probability distribution of EXPUCB changes more drastically than that of B-EXPUCB, this is because it is easier for EXPUCB to avoid the adaptive attacker than B-EXPUCB.  Although the LinUCB-random also randomizes among the paths, it fails to learn the best paths suitable for the ResNet50 tasks, resulting in a lower reward. 

\vspace{-5 pt}
\begin{figure}[th]
    \centering
        \subfigure[Accumulated rewards]{
        \begin{minipage}[b]{0.49\linewidth}
            \includegraphics[width=0.95\textwidth]{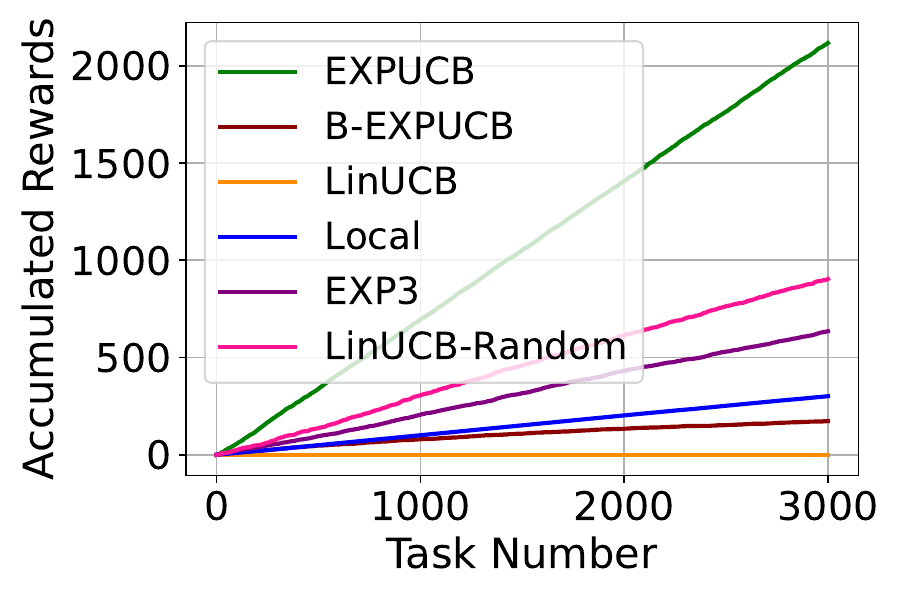}
        \end{minipage}
        \label{fig:rewardAdaptiveAttacker}
    }
    \hspace{-6mm}
    \subfigure[Accumulated switching cost]{
        \begin{minipage}[b]{0.49\linewidth}
        \includegraphics[width=0.95\textwidth]{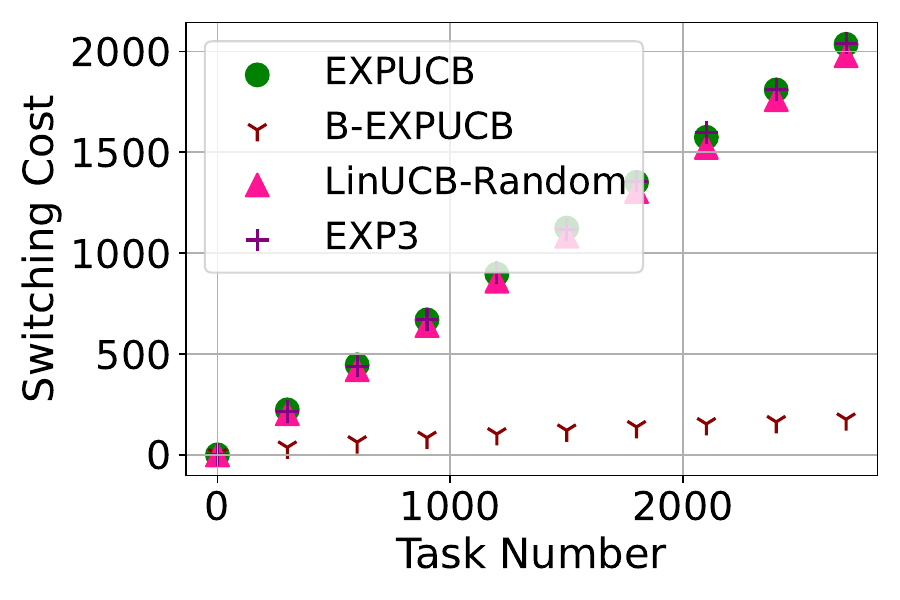}
        \end{minipage}
    \label{fig:switchingcostAdaptiveAttacker}
    }
    \vspace{-5 pt}
    \caption{Performance of B-EXPUCB under the adaptive attacking strategy. (ResNet50-only tasks).}
    \label{fig:AdaptiveAttacker}
    \vspace{-10 pt}
\end{figure}

\vspace{-5 pt}

\section{Conclusion}
In this paper, we address the complex challenge of joint path selection and DNN layer assignment for collaborative DNN inference, a critical component in optimizing distributed neural network operations. We derive structural results concerning the DNN layer assignment, significantly reducing the decision space. These findings provide profound insights into the optimal DNN layer assignment, assuming full knowledge of the network parameters, thereby enhancing the efficiency of DNN inference. In scenarios where node processing speeds and link transmission rates remain unknown, we introduce a novel bandits algorithm named B-EXPUCB. This algorithm is designed to incrementally learn the most effective path and DNN layer assignments over time. We establish sublinear regret bounds for B-EXPUCB, demonstrating its asymptotically optimal learning performance. Our findings not only underscore the robustness of B-EXPUCB in uncertain environments but also have significant implications for the adaptation and evolution of distributed DNN inference systems in dynamic settings.

\bibliographystyle{IEEEtran}
\bibliography{reference}

\clearpage
\appendix
\section*{Proof of Lemma 1}
\begin{proof}
This lemma follows Theorem 2 in \cite{abbasi2011improved} by considering the sub-sequence of rounds in which group $g$ is selected by the learner and not attacked by the adversary. We have for all $g$ and $t$ with probability at least $1-\delta$,
\begin{align}
    \|\hat{\theta}^t_g - \theta_g\|_{V^t_g} \leq \alpha^t.
\end{align}
Then \eqref{error_bound} holds by the Cauchy-Schwarz inequality. 
\end{proof}

\section*{Proof of Lemma 2}
\begin{proof}
Consider the sub-sequence of rounds in which group $g$ is selected by the learner and not attacked by the adversary, we have
\begin{align}
    &\sum_{\tau=s}^{S} \sqrt{(x^s)^\top (V^{s-1}_g)^{-1} x^s} \\
    \leq &\sqrt{S \sum_{s=1}^S (x^s)^\top (V^{s-1}_g)^{-1} x^s} \nonumber\\
    \leq &\sqrt{S 2d \left(\log(\lambda + \frac{S}{d}) - \log \lambda \right)} \nonumber\\
    \leq & \sqrt{T 2d \left(\log(\lambda + \frac{T}{d}) - \log \lambda \right)}, \nonumber
\end{align}
where the first inequality is due to the Jensen’s inequality, the second inequality is due to Lemma 11 in \cite{abbasi2011improved}, and the last inequality is because the length of the sub-sequence $S$ is smaller than $T$. Further noticing that
\begin{align}
    \alpha^t \leq \alpha^T = \sqrt{\lambda} + \sigma \sqrt{d\log\frac{1+T/\lambda}{\delta}}
\end{align}
yields the desired bound. 
\end{proof}

\section*{Proof of Theorem 1}
\begin{proof}
Denote $w^t(g) = \exp(\eta R^{t-1}_g)$, $W^t = \sum_{g'=1}^G w^t(g')$ and $I^t(g) = \frac{\textbf{1}\{g^t = g\}}{P^t(g)}$. We first observe that $W^{T_{N+1}}$ can be lowered bounded with probability at least $1-\delta$ as follows
{\allowdisplaybreaks
\begin{align}
    &\log\left(\frac{W^{T_{N+1}}}{W^1}\right)\label{wt+1_wt}
    \geq \log\left(\frac{\exp(\eta\sum_{t=1}^{T_{N+1}} I^t(\gamma)r^t/B)}{W^1}\right) \\
    = & \eta \frac{\sum_{t=1}^{T_{N+1}}I^t(\gamma) r^t}{B} - \log G \nonumber\\
    = & \eta \frac{\sum_{t=1}^{T_{N+1}}I^t(\gamma) a^t(\gamma)(\theta_\gamma^\top x^t + n^t)}{B} - \log G \nonumber\\
    \geq & \eta\frac{\sum_{t=1}^{T_{N+1}} I^t(\gamma) a^t(\gamma)(\theta_\gamma^\top \xi^t_\gamma + n^t)}{B} - \log G \nonumber\\
    &- \eta\frac{\sum_{t=1}^{T_{N+1}} I^t(\gamma) a^t(\gamma) 2\alpha^t\sqrt{(x^t)^\top (V^{t-1}_{\gamma})^{-1}x^t}}{B}, \nonumber
\end{align}
where the first inequality uses $W^{T_{N+1}} \geq w^{T_{N+1}}(\gamma)$, the second equality uses $W^1 = G$, the third equality uses the definition of $r^t$, and the last inequality is derived based on the arm selection rule \eqref{arm-estimate}. Specifically, we have the following lower bound on $r^t$ with probability at least $1-\delta$ for all $t$ such that $I^t(\gamma) a^t(\gamma) = 1$,
}
{\allowdisplaybreaks
\begin{align}
    r^t =& \theta_\gamma^\top x^t + n^t \label{reward_upper_bound}\\
    \geq &  (\hat{\theta}^{t-1}_\gamma)^\top x^t - \alpha^t\sqrt{(x^t)^\top (V^{t-1}_\gamma)^{-1} x^t} + n^t \nonumber\\
    \geq & (\hat{\theta}^{t-1}_\gamma)^\top \xi^t_\gamma + \alpha^t \sqrt{(\xi^t_\gamma)^\top (V^{t-1}_\gamma)^{-1}\xi^t_\gamma} \nonumber\\
    & -2\alpha^t\sqrt{(x^t)^\top (V^{t-1}_\gamma)^{-1} x^t} + n^t \nonumber\\
    \geq &\theta_\gamma^\top \xi^t_\gamma -2\alpha^t\sqrt{(x^t)^\top (V^{t-1}_\gamma)^{-1} x^t} + n^t, \nonumber
\end{align}
where the first equality is the definition of $r^t$, the second/fourth inequality uses the lower/upper confidence bound in Lemma 1 which holds with probability at least $1-\delta$, and the third inequality uses the arm selection rule \eqref{arm-estimate}.}

On the other hand, we have the following upper-bound
{\allowdisplaybreaks
\begin{align}
    &\log\left(\frac{W^{T_{n+1}}}{W^{T_n}}\right)\\
    =&\log\left(\sum_{g=1}^G \frac{\exp(\eta\sum_{q=1}^{T_{n+1}-1} I^q(g) r^q/B)}{W^{T_n}}\right) \nonumber\\
    =&\log\left(\sum_{g=1}^G \frac{\exp(\eta\sum_{q=1}^{T_{n}-1} I^q(g) r^q/B)}{W^{T_n}}\exp(\sum_{q=T_n}^{T_{n+1}-1}\eta I^q(g) r^q/B)\right) \nonumber\\
    =&\log\left(\sum_{g=1}^G \frac{P^{T_n}(g) - \frac{\beta}{G}}{1-\beta}\exp(\sum_{q=T_n}^{T_{n+1}-1}\eta I^q(g) r^q/B)\right) \nonumber\\
    \leq & \log\Bigg(\sum_{g=1}^G \frac{P^{T_n}(g) - \frac{\beta}{G}}{1-\beta}(1 + \frac{\sum_{q=T_n}^{T_{n+1}-1}\eta I^q(g) r^q}{B} \nonumber \\&+ (\frac{\sum_{q=T_n}^{T_{n+1}-1}\eta I^q(g) r^q}{B})^2)\Bigg) \nonumber\\
    \leq & \Bigg(\sum_{g=1}^G \frac{P^{T_n}(g) - \frac{\beta}{G}}{1-\beta}(1 + \frac{\sum_{q=T_n}^{T_{n+1}-1}\eta I^q(g) r^q}{B} \nonumber \\&+ (\frac{\sum_{q=T_n}^{T_{n+1}-1}\eta I^q(g) r^q}{B})^2)\Bigg) - 1 \nonumber\\
    \leq & \sum_{g=1}^G \frac{P^{T_n}(g)}{1-\beta}\Bigg(\sum_{q=T_n}^{T_{n+1}-1}\eta I^q(g) r^q/B + (\sum_{q=T_n}^{T_{n+1}-1}\eta I^q(g) r^q/B)^2\Bigg) \nonumber\\
    &- \frac{\eta \beta}{G(1-\beta)B}\sum_{g=1}^G\sum_{q=T_n}^{T_{n+1}-1}\eta I^q(g) r^q, \nonumber
\end{align}
where the first equality uses the definition of $W^{T_{n+1}}$, the second equality breaks the sum into two parts, the third equality uses the definition of the sampling distribution $P^t$, the forth inequality uses $e^z \leq 1 + z + z^2, \forall z \leq 1$, the fifth inequality uses $\log z \leq z -1, \forall z \geq 0$, and the last inequality holds by canceling out terms and realizing that $-\sum_{g=1}^G(\sum_{q=T_n}^{T_{n+1}-1}\eta I^q(g) r^q/B)^2 \leq 0$. 
Noticing that $\sum_{n=1}^N \log\frac{W^{T_{n+1}}}{W^{T_n}} = \log \frac{W^{T+1}}{W^1}$, we can sum both sides for $n = 1, ..., N$ and get

\begin{align}
    \log\frac{W^{T_{N+1}}}{W^1} &\le \sum_{t=1}^{T_{N+1} - 1} (\sum_{g=1}^G \frac{P^t(g)}{1-\beta}(\eta I^t(g) r^t/B ) \\+& \sum_{n=1}^N\sum_{g=1}^G\Bigg(\frac{P^{T_n}(g)}{1-\beta}(\sum_{q=T_n}^{T_{n+1}-1}\eta I^q(g) r^q/B)^2  \nonumber
    \\-& \frac{\eta \beta}{G(1-\beta)B} I^t(g) r^t \Bigg)\nonumber
\end{align}

Compare with the lower bound in \eqref{wt+1_wt} and obtain
{\allowdisplaybreaks
\begin{align}
    &\frac{\eta\sum_{t=1}^{T_{N+1}-1} I^t(\gamma) a^t(\gamma)( \theta_\gamma^\top \xi^t_\gamma + n^t)}{B} - \log G\label{confidence_upper_bound} \\-& \frac{\eta\sum_{t=1}^{T_{N+1} -1 } I^t(\gamma) a^t(\gamma) 2\alpha^t\sqrt{(x^t)^\top (V^{t-1}_{\gamma})^{-1}x^t}}{B} \nonumber\\
    \leq & \sum_{t=1}^{T_{n+1} - 1} (\sum_{g=1}^G \frac{P^t(g)}{(1-\beta)B}(\eta I^t(g) r^t ) \nonumber \\+& \sum_{n=1}^N\sum_{g=1}^G\left(\frac{P^{T_n}(g)}{1-\beta}(\sum_{q=T_n}^{T_{n+1}-1}\eta I^q(g) r^q/B)^2 - \frac{\eta \beta}{GB(1-\beta)} I^t(g) r^t \right) \nonumber
\end{align}}
Reordering and multiplying both sides by $\frac{(1-\beta)B}{\eta}$ gives
\begin{align}
    &\sum_{t=1}^{T_{N+1} -1} \left(I^t(\gamma) a^t(\gamma)( \theta_\gamma^\top \xi^t_\gamma + n^t) - \sum_{g=1}^G\textbf{1}\{g^t = g\} r^t\right)\\
    \leq & \frac{(1-\beta)B}{\eta}\log G + \sum_{n=1}^N\sum_{g=1}^G\frac{P^{T_n}(g)}{B^2\eta}(\sum_{q=T_n}^{T_{N+1}-1}\eta I^q(g) r^q)^2  \nonumber\\
    & + \beta \sum_{t=1}^{T_{N+1}-1} \left(I^t(\gamma) a^t(\gamma)(\theta_\gamma^\top \xi^t_\gamma + n^t) - \frac{1}{G} \sum_{g=1}^G I^t(g) r^t\right) \nonumber\\
    & + (1-\beta) \sum_{t=1}^{T_{N+1}-1} I^t(\gamma) a^t(\gamma) 2\alpha^t\sqrt{(x^t)^\top (V^{t-1}_{\gamma})^{-1}x^t}. \nonumber
\end{align}

Besides, consider the regret of  timeslots in the last block from $T_{N+1}$ to $T$,

\begin{align*}
    &\mathbb{E}[\sum_{t={T_{N+1}}}^{T} r^t(\gamma, \xi^t_\gamma) - \sum_{t=1}^T r^t(g^t, x^t)] 
    \le B \label{last_round_regret}, 
\end{align*}
where this inequality holds by the maximal total regret between rounds ${T_{N+1}}$ and $T$ is at most B.

Now, consider the regret, which can be written alternatively as follows
\begin{align}
  &\reg(T) = \mathbb{E}[\sum_{t=1}^T r^t(\gamma, \xi^t_\gamma) - \sum_{t=1}^T r^t(g^t, x^t)]\\
   =&\mathbb{E}\left[\sum_{t=1}^T \left(I^t(\gamma) a^t(\gamma)(\theta_\gamma^\top \xi^t_\gamma + n^t) - \sum_{g=1}^G \textbf{1}\{g^t = g\} r^t\right)\right]\label{reg_def}, \nonumber
\end{align}
where the second equality uses
\begin{align}
    &\mathbb{E}[r^t(\gamma, \xi^t_\gamma)] = \mathbb{E}[\theta_\gamma^\top \xi^t_\gamma + n^t]\\
    &\mathbb{E}[I^t(\gamma)] = 1.
\end{align}
Plugging the bound into \eqref{confidence_upper_bound} and \eqref{last_round_regret} yields
\begin{align} 
    &\reg(T)\\
    \leq &\frac{(1-\beta)B}{\eta}\log G + \eta GN + \beta GT + B\nonumber\\
    &+\frac{1-\beta}{\beta} \mathbb{E}\left[\sum_{t=1}^T \textbf{1}\{g^t = \gamma\}a^t(\gamma)2\alpha^t \sqrt{(x^t)^\top (V^{t-1}_\gamma)^{-1} x^t} \right]  \nonumber\\
    \leq&\frac{B}{\eta}\log G + \frac{\eta GT}{B} + \beta T + B \nonumber\\
    +&\frac{2}{\beta}\left(\sqrt{\lambda} + \sigma \sqrt{d\log\frac{\lambda+T}{\delta\lambda}}\right)\sqrt{2T d \left(\log(\lambda + \frac{T}{d}) - \log \lambda \right)}, \nonumber
\end{align}
where the last inequality is due to Lemma 2.}

Finally, by setting $\beta = T^{-1/4}\sqrt{\log(T)}$, $\eta = T^{1/6}$, and $B = T^{2/3}$, we have $R_T = O(T^{3/4}\sqrt{\log(T)})$.

Note that the number of switches is bounded by the number of blocks, i.e., $\Gamma + 1$. Thus, the cumulative switching regret satisfies
\begin{align*}
    S_T = N + 1 =  \lceil T/B \rceil =  O(T^\frac{1}{3}).
\end{align*}

Therefore, the total regret is
\begin{align}
    \reg(T) = R_T + S_T \leq O(T^{3/4}\sqrt{\log(T)})
\end{align}    
\end{proof}

\end{document}